%% file: arxiv_main.tex
\documentclass[wcp]{jmlr}

\usepackage[utf8]{inputenc}
\usepackage[T1]{fontenc}
\usepackage{mathtools}
\usepackage{bm}
\usepackage{dsfont}
\usepackage{booktabs}
\usepackage{xcolor}
\usepackage{longtable}
\usepackage{float}
\usepackage{lineno}
\hypersetup{
  pageanchor=false,
  hypertexnames=false,
  pdftitle={Exact Selective Inference for Spectral Clustering via Piecewise-Constant Graph Perturbation},
  pdfauthor={Naotoshi Nakamura},
  pdfkeywords={selective inference, spectral clustering, post-clustering inference, graph homotopy, truncated pivots}
}

\makeatletter
\let \@jmlrpages \@empty
\makeatother

\newtheorem{assumption}[theorem]{Assumption}
\newtheorem{suppprop}{Supplementary Proposition}

\newcommand{\R}{\mathbb{R}}
\newcommand{\E}{\mathbb{E}}
\newcommand{\Prob}{\mathbb{P}}
\newcommand{\X}{\mathbf{X}}
\newcommand{\x}{\mathbf{x}}
\newcommand{\bmu}{\bm{\mu}}
\newcommand{\nuv}{\bm{\nu}}
\newcommand{\Cc}{\mathcal{C}}
\newcommand{\Spec}{\mathcal{C}_{\mathrm{spec}}}
\newcommand{\Cab}{\{C_a,C_b\}}
\newcommand{\dir}{\mathrm{dir}}
\newcommand{\Pp}{\Pi_{\nuv}^{\perp}}
\newcommand{\Acal}{\mathcal{A}}

\jmlrvolume{353,}
\jmlryear{2026}
\jmlrworkshop{ACML 2026}
\title[Exact SI for Spectral Clustering]{Exact Selective Inference for Spectral Clustering via Piecewise-Constant Graph Perturbation}
\author{\Name{Naotoshi Nakamura} \Email{nakamura.nao.ms@yokohama-cu.ac.jp}\\
\addr{Graduate School of Data Science, Yokohama City University, Yokohama, Japan}}
\editors{Andy Song, Bo Han and Sarah Erfani}

\begin{document}

\makeatletter
\let \@jmlrpages \@empty
\makeatother

\maketitle

\begin{abstract}
Cluster labels are often used twice: first to discover groups, and then to test
whether the discovered groups differ. This double use of the data makes ordinary
two-sample $p$-values anti-conservative. Selective inference fixes this by
conditioning on the clustering event, but for spectral clustering the event
appears difficult because the algorithm combines an affinity map, an
eigendecomposition, and a rounding step.

We show that, for spectral clustering based on unweighted $\varepsilon$- or
$k$-nearest-neighbor graphs, the obstruction is the affinity construction rather
than the eigendecomposition. Along the one-dimensional perturbation used by the
truncated-$\chi$ selective test, graph changes occur only at finitely many real
roots of data-dependent quadratic distance comparisons. Between these
breakpoints, the graph, Laplacian spectrum, embedding, rounding, and reported
cluster pair are deterministic functions of a finite graph state. Consequently,
in real arithmetic and under a known-scale spherical Gaussian model, the
selection set is an exact finite union of intervals, yielding finite-sample
selective $p$-values for selected spectral-cluster pairs. The same argument gives
feature-wise truncated-normal tests. We also record direct extensions to known
feature covariance, eigengap-selected $K$, and auxiliary-coordinate graphs. We
validate null and partial-null calibration, operational power, numerical
sensitivity, and model-scope diagnostics, and give a working-model PBMC
algorithmic illustration. Smooth kernels such as RBF affinities fall outside the
finite-state guarantee.
\end{abstract}

\section{Introduction}
Clustering is routinely followed by a confirmatory step: having obtained
clusters, the analyst asks whether two of them are truly different, for
example whether their feature means differ. The naive approach, namely a two-sample
test between the estimated clusters, is invalid, because the clusters were
chosen by the data precisely to be separated, so the test rejects far too often
even when no structure is present (the ``double-dipping'' problem~\citep{ball2019double}). 
Selective inference resolves this by conditioning on the event that the algorithm returned
the clusters of interest~\citep{lee2016exact,gao2024selective}.

For hierarchical clustering with squared-Euclidean Ward linkage,
\citet{gao2024selective} derived an \emph{exact} selective test: under a spherical
Gaussian model, the difference-in-means statistic follows a $\chi$ distribution
truncated to the set of perturbations that preserve the clusters, and that set
is an analytically computable union of intervals. \citet{chen2023selective}
extended the program to $k$-means by conditioning on the iterates of Lloyd's
algorithm. Subsequent work has extended post-clustering selective inference to
convex clustering~\citep{bachoc2023convex}, dependent observations and
features~\citep{gonzalezdelgado2023dependency}, and functional or longitudinal
data~\citep{zhu2024functional}; see also the recent review of conditional
selective inference by \citet{neufeld2026review}. A parallel line based on
parametric programming
\citep{leduy2022parametric} computes the same conditioning set by line search
and has recently been assembled into a framework for multi-stage clustering
\emph{pipelines}~\citep{miyata2026pipeline}, whose component library covers
distance-based outlier detection, variance/correlation feature selection,
$k$-means and DBSCAN. A method-agnostic alternative, data
thinning~\citep{neufeld2024thinning,dharamshi2025generalized}, splits the data
into independent parts and avoids conditioning by testing clusters selected on
an independent noisy copy. This is valid, but it targets that randomized
selection copy rather than the deterministic clustering returned on the observed
data, which is our target.

Conspicuously absent from all of these is \emph{spectral}
clustering~\citep{ng2002spectral,vonluxburg2007tutorial}, despite its
ubiquity. Graph-based representations and decompositions are also central in
genomic interaction analysis, including graph-cover formulations for Hi--C data
and comparative evaluation of topological-domain callers on RNA-associated
interactomes~\citep{sefer2021hic,alaygut2026tad}. Spectral clustering composes three operations: an affinity map, an
eigendecomposition of a graph Laplacian, and a rounding step ($k$-means on the
spectral embedding); and the eigendecomposition is the step that does not
admit the polyhedral / linear-or-quadratic characterization of the
selection event on which existing exact post-clustering inference
relies~\citep{gao2024selective,bachoc2023convex,leduy2022parametric}. It is
therefore natural to suspect that the eigendecomposition is the obstruction. Our
central observation is that \textbf{it is not}; the affinity construction is. We
make this precise and instantiate an exact-in-real-arithmetic selective test. Many spectral
clustering pipelines construct an unweighted $\varepsilon$- or
$k$-nearest-neighbor graph before forming a graph Laplacian; our theorem targets
exactly this graph-based regime, while smooth weighted kernels require separate
analysis.

\paragraph{Contributions.}
\begin{enumerate}
\item \textbf{Finite-state theorem.} We prove that any polynomial finite-state
graph selector yields a finite-union selection set for deterministic spectral
clustering (Section~\ref{sec:dichotomy}).
\item \textbf{Neighborhood graphs.} Unweighted $\varepsilon$-graphs and
OR-symmetrized $k$-NN graphs satisfy the theorem with quadratic predicates and
quadratic-root breakpoints.
\item \textbf{Selective tests.} Under the stated known-scale spherical Gaussian
model, we derive finite-sample exact truncated-$\chi$ tests in real arithmetic
for selected spectral-cluster pairs, with deterministic gap, rounding, and
target-selection conventions.
\item \textbf{Feature-wise inference.} We derive truncated-normal tests for
marker or driver features after spectral clustering.

\item \textbf{Direct extensions and scope.} We record short extensions for known
feature covariance, eigengap-selected $K$, and auxiliary-coordinate affinities,
while keeping the main theorem focused on the fixed-$K$ neighborhood-graph case.
\item \textbf{Empirical validation and scope.} We report global- and
partial-null calibration, operational power and usability, numerical-sensitivity
and misspecification diagnostics, scaling behavior, and a working-model PBMC
illustration (Section~\ref{sec:experiments}). To our knowledge, this is the
first finite-sample exact selective-inference procedure, at the theorem level,
for spectral clustering with unweighted neighborhood-graph affinities.
\end{enumerate}

\section{Setup and selective test}\label{sec:setup}
\paragraph{Model.}
Let $\X\in\R^{n\times d}$ have independent rows
$\x_i\sim N(\bmu_i,\sigma^2 I_d)$ with $\sigma$ \emph{known}. A clustering
algorithm is a map $\Cc:\R^{n\times d}\to\Pi_n$ to partitions of
$\{1,\dots,n\}$; write $\Spec$ for spectral clustering. For two clusters
$C_a,C_b\in\Cc(\X)$ we test
\begin{equation}
H_0:\ \bar\bmu_{C_a}=\bar\bmu_{C_b},
\qquad
\bar\bmu_{C}:=\tfrac{1}{|C|}\textstyle\sum_{i\in C}\bmu_i .
\end{equation}

\paragraph{Contrast and statistic.}
Define $\nuv\in\R^n$ by
$\nu_i = \mathds{1}(i\in C_a)/|C_a| - \mathds{1}(i\in C_b)/|C_b|$, so that
$\nuv^\top\X=\bar\x_{C_a}-\bar\x_{C_b}\in\R^d$ and
$\E[\nuv^\top\X]=\bar\bmu_{C_a}-\bar\bmu_{C_b}$. The test statistic is
$T(\X)=\|\nuv^\top\X\|_2$. Under $H_0$,
$\nuv^\top\X\sim N(0,\sigma^2\|\nuv\|^2 I_d)$ and hence
\begin{equation}\label{eq:chi}
T(\X)\,/\,(\sigma\|\nuv\|)\ \sim\ \chi_d .
\end{equation}
Here $\chi_d$ denotes the distribution of $\|G\|_2$ for
$G\sim N(0,I_d)$; it is a chi, not a chi-squared, distribution.
Note $\|\nuv\|^2=1/|C_a|+1/|C_b|$, and $\nuv$ is supported only on
$C_a\cup C_b$: points outside the tested pair (``spectators'') do not enter the
contrast, though they still shape the graph.

\paragraph{Perturbation path.}
Let $\pi:=\dir(\nuv^\top\X)=\nuv^\top\X/T(\X)$. Define, for $\phi\ge0$,
\begin{equation}\label{eq:path}
\X'(\phi)\ :=\ \X+\frac{\phi-T(\X)}{\|\nuv\|^2}\,\nuv\,\pi^\top ,
\end{equation}
so that $\nuv^\top\X'(\phi)=\phi\,\pi$, the orthogonal nuisance
$\Pp\X$ and direction $\pi$ are preserved, and $\X'(T(\X))=\X$. Crucially,
\eqref{eq:path} is \emph{affine in $\phi$}: row $i$ is $p_i+\phi\,q_i$ with
$p_i=\x_i-(\nu_i/\|\nuv\|^2)\,\nuv^\top\X$ and $q_i=(\nu_i/\|\nuv\|^2)\,\pi$;
spectator rows have $q_i=0$.

\paragraph{Target-selection rule.}
The tested pair may itself be chosen after clustering, so we make that rule part
of the selection event. Let $\Acal$ be a deterministic target-selection rule that
maps the clustering output (and any deterministic diagnostics such as cluster
sizes) to a set of unordered cluster pairs to be tested. For example,
$\Acal_{\mathrm{app}}(\Cc)$ returns all unordered pairs of clusters in $\Cc$,
whereas $\Acal_{\mathrm{2lg}}(\Cc)$ returns only the pair consisting of the two
largest clusters, with a fixed tie-breaking rule. The observed target is the
unordered pair $\Cab$ of index sets. Once the deterministic validity conventions
of Section~\ref{sec:deterministic_pipeline} are fixed, the matched selection
event is that the spectral pipeline is valid and $\Cab\in\Acal(\Spec(\X))$.
If the reported hypothesis is simply this observed pair, $\Acal_{\mathrm{app}}$
is the Gao--Bien--Witten ``pair appears'' event; if the analysis protocol says
to test the two largest clusters, $\Acal_{\mathrm{2lg}}$ is the correct event to
condition on. More generally, any pre-specified rule that is a deterministic
function of the finite graph state and pipeline output is covered without new
graph roots. A rule using additional raw-data summaries requires its own
selection decision to be added, and the finite-state conclusion persists only
when that added decision is itself piecewise constant with finitely many
tractable boundaries.

\begin{table}[!ht]
\centering
\caption{Conditioning levels used in the paper. The valid event must match the
reported target; conditioning on more information is valid but can be
conservative.}
\label{tab:conditioning-hierarchy}
\small
\resizebox{\textwidth}{!}{%
\begin{tabular}{lll}
\toprule
Level & Event fixed along the path & Role \\
\midrule
Graph-conditioned & realized graph state & cheapest, most conservative \\
Full partition & exact spectral partition & intermediate diagnostic \\
Matched rule & $\Cab\in\Acal(\Cc(\cdot))$ for the reporting rule & primary test \\
Pair appearance & $\Cab$ appears among cluster pairs & valid when the pair is user chosen \\
\bottomrule
\end{tabular}
}
\end{table}
\FloatBarrier

\paragraph{Selective $p$-value.}
Conditioning on the matched selection event together with the nuisances
$(\Pp\X,\pi)$ reduces the test to a one-dimensional truncated $\chi$
problem~\citep{gao2024selective}. With
\begin{equation}\label{eq:S}
S_{\Acal}\ :=\ \{\phi\ge0:\ \Spec(\X'(\phi))\ \text{is valid and}\
\Cab\in\Acal(\Spec(\X'(\phi)))\},
\end{equation}
the selective $p$-value is
\begin{equation}\label{eq:pval}
p\ =\ \frac{\Prob\big(\sigma\|\nuv\|\,\chi_d\ge T(\x^{\mathrm{obs}}),\ \sigma\|\nuv\|\,\chi_d\in S_{\Acal}\big)}
{\Prob\big(\sigma\|\nuv\|\,\chi_d\in S_{\Acal}\big)} .
\end{equation}
If $S_{\Acal}$ is a union of intervals $\bigcup_m[\ell_m,u_m]$, \eqref{eq:pval}
is a finite sum of $\chi_d$-CDF differences. The entire problem thus reduces to
computing $S_{\Acal}$ for the actual target-selection rule.

The spherical model $\x_i\sim N(\bmu_i,\sigma^2 I_d)$ can be relaxed when the
feature covariance is known: one applies the entire pipeline after whitening, as
summarized with the direct extensions in Section~\ref{sec:extensions}.

\section{The tractability dichotomy}\label{sec:dichotomy}
Because \eqref{eq:path} is affine, every pairwise squared distance along the
path is a scalar quadratic in $\phi$:
\begin{equation}\label{eq:quad}
d_{ij}(\phi)^2=\|(p_i-p_j)+\phi(q_i-q_j)\|^2
=\alpha_{ij}\phi^2+\beta_{ij}\phi+\gamma_{ij},
\end{equation}
with $\alpha_{ij}=\|q_i-q_j\|^2$,
$\beta_{ij}=2(p_i-p_j)^\top(q_i-q_j)$, $\gamma_{ij}=\|p_i-p_j\|^2$. What happens
next depends sharply on the affinity.

\begin{definition}[Finite-state selector]\label{def:finite}
Fix the nuisance variables defining the affine path \eqref{eq:path}. A graph
construction is polynomial finite-state along this path if finitely many real
polynomials $h_1,\dots,h_M$ and a deterministic map $\Gamma$ satisfy
\[
G(\phi)=\Gamma\!\big(\operatorname{sign}h_1(\phi),\dots,
\operatorname{sign}h_M(\phi)\big)
\]
for all $\phi$ outside the roots of the $h_m$'s. If each $h_m$ has degree at most
$r$, we call the selector degree-$r$.
\end{definition}

\begin{lemma}[Fixed graph states]\label{lem:detinterval}
Fix a deterministic graph-to-clustering pipeline with its invalid-state
convention, and fix a target-selection rule $\Acal$. If $G(\phi)=G_0$ on an
interval $I$, then the clustering output and the decision
\[
\mathds{1}\{\Cab\in\Acal(\Spec(G(\phi)))\}
\]
are constant on $I$, where the indicator is defined to be zero whenever the
pipeline declares the graph state invalid.
\end{lemma}

\begin{theorem}[Finite-state graph selector theorem]\label{thm:finite}
Suppose the graph along the selective path is a degree-$r$ polynomial
finite-state selector. Suppose further that the embedding, spectral-gap check,
rounding map, and target-selection rule $\Acal$ are deterministic functions of
the graph state. Then the selection indicator
\[
\phi\ \longmapsto\ \mathds{1}\{\Cab\in\Acal(\Cc(\X'(\phi)))\}
\]
is constant on every connected component of
$[0,\infty)\setminus\bigcup_{m=1}^M\{\phi:h_m(\phi)=0\}$. Consequently
$S_{\Acal}$ in \eqref{eq:S} is, up to endpoints of measure zero, a finite union
of intervals whose finite endpoints are roots of the $h_m$'s. In particular, if
the $h_m$ have algebraic coefficients and degree at most $r$, the endpoints are
algebraic numbers.
\end{theorem}

\begin{corollary}[Neighborhood graphs]\label{cor:neighborhood}
Unweighted $\varepsilon$-graphs and OR-symmetrized $k$-NN graphs are degree-$2$
finite-state selectors along \eqref{eq:path}. Their candidate breakpoints are
contained in the real roots of
\begin{itemize}
\item[(i)] $d_{ij}(\phi)^2=\varepsilon^2$ over all pairs $(i,j)$
($\varepsilon$-graph; $O(n^2)$ candidate roots), or
\item[(ii)] $d_{ij}(\phi)^2=d_{im}(\phi)^2$ over all triples $(i,j,m)$
($k$-NN graph; $O(n^3)$ candidate roots).
\end{itemize}
Each equation is quadratic in $\phi$, so, in real arithmetic,
$S_{\Acal}$ is an exact finite union of intervals whose finite endpoints are
among the real roots of these data-dependent quadratics, for any deterministic
spectral pipeline and deterministic target-selection rule $\Acal$.
\end{corollary}
Proofs of Lemma~\ref{lem:detinterval}, Theorem~\ref{thm:finite}, and
Corollary~\ref{cor:neighborhood} are given in the supplement.

\begin{remark}[The eigendecomposition is not the obstruction]
Theorem~\ref{thm:finite} never differentiates eigenvectors in $\phi$: on each
finite-state interval the Laplacian is a \emph{fixed} matrix, so its spectrum is
fixed. All $\phi$-dependence is combinatorial, living in the graph selector and
in the deterministic target-selection rule. This is why unweighted graphs are
the tractable case.
\end{remark}

\begin{remark}[RBF affinities]
For a Gaussian/RBF affinity
$w_{ij}(\phi)=\exp(-d_{ij}(\phi)^2/2h^2)$, $W(\phi)$ is real-analytic rather
than finite-state. Away from eigenvalue crossings and rounding ties,
eigenvectors vary analytically by Rellich's theorem. The endpoints of the
selection set are then zeros of analytic, generally transcendental functions,
not roots of finitely many graph-state polynomials. A line-search method may
localize such endpoints numerically under suitable bracketing and tie-free
conditions, but the finite quadratic-root interval representation of
Corollary~\ref{cor:neighborhood} is no longer supplied by our finite-state
argument.
\end{remark}

\section{A finite-state spectral selective test}\label{sec:method}
We instantiate the construction with a single code path parameterized by the
graph type (unweighted $\varepsilon$- or $k$-NN) and the Laplacian
(unnormalized $L=D-W$, symmetric-normalized $L_{\mathrm{sym}}=I-D^{-1/2}WD^{-1/2}$
with row-normalized embedding \`a la Ng--Jordan--Weiss, or random-walk
$L_{\mathrm{rw}}=I-D^{-1}W$ \`a la Shi--Malik), with $k$-means rounding for
general $K$, testing a chosen pair of clusters. All variants share the same
breakpoint set (Corollary~\ref{cor:neighborhood}) and the same gap condition
below, since $L_{\mathrm{rw}}$ and $L_{\mathrm{sym}}$ have identical eigenvalues.

\subsection{Deterministic spectral pipeline and degeneracy conventions}
\label{sec:deterministic_pipeline}
For the exact selection event, spectral clustering must be a single-valued map
from a graph state to either a partition or an invalid symbol. We use the
following convention throughout.

\paragraph{Graph construction.}
The $\varepsilon$-graph has
$W_{ij}=\mathds{1}\{d_{ij}^2\le\varepsilon^2\}$ for $i\ne j$ and zero diagonal,
so the endpoint convention is closed. For the $k$-NN graph, self-neighbors are
excluded and node $i$ orders candidates by the lexicographic key
$(d_{ij}^2,j)$; $N_k(i)$ is the first $k$ indices in that order. The undirected
graph is the OR-symmetrization
$W_{ij}=\mathds{1}\{j\in N_k(i)\ \text{or}\ i\in N_k(j)\}$. Mutual $k$-NN is not
used in the reported experiments.

\paragraph{Degrees, Laplacians, and zero rows.}
Let $d_i=\sum_j W_{ij}$. For normalized variants we set
$d_i^{-1}=d_i^{-1/2}=0$ when $d_i=0$; thus isolated vertices contribute zero
rows to the normalized adjacency. We then form
$L=D-W$,
$L_{\mathrm{sym}}=I-D^{-1/2}WD^{-1/2}$, or, for the random-walk embedding, the
equivalent symmetric generalized problem obtained from $L_{\mathrm{sym}}$ and
mapped back by $D^{-1/2}$. Zero-degree states are not random choices; they are
handled by this convention and may still be excluded by the gap check below.

\paragraph{Gap and bottom-$K$ representation.}
A graph state is declared invalid unless
$\lambda_K(L)<\lambda_{K+1}(L)$ for the chosen Laplacian eigenvalues in
nondecreasing order. Repeated eigenvalues inside the bottom-$K$ subspace are
allowed. To remove basis ambiguity, the mathematical embedding is defined from
the bottom-$K$ spectral projector by the following deterministic procedure:
\begin{enumerate}
\item Scan the coordinate vectors $e_1,\dots,e_n$ in index order.
\item Project each vector onto the bottom-$K$ invariant subspace.
\item Orthogonalize the projection against all previously accepted vectors.
\item Append the normalized residual whenever it is nonzero.
\item Choose the sign so the first nonzero coordinate is positive.
\end{enumerate}
The resulting canonical orthonormal basis is a deterministic representation of
the selected invariant subspace; away from degeneracies it agrees with the usual
spectral-clustering target up to rounding. In floating-point experiments we use
the same convention up to fixed numerical tolerances, declaring a graph invalid
when the observed gap is below tolerance.

\paragraph{Embedding, rounding, and labels.}
For $L_{\mathrm{sym}}$ we row-normalize the $n\times K$ embedding; a zero
spectral row remains zero. For $L_{\mathrm{rw}}$ and $L$ we use the unnormalized
rows. Rounding is deterministic Lloyd $k$-means: a fixed pseudorandom seed gives
the initialization, center-distance ties are assigned to the smallest center
index, center updates are performed in label order, empty clusters retain their
previous center, and the iteration stops when labels no longer change or after
a fixed maximum number of iterations. The inferential target is expressed as an
unordered pair of cluster index sets, so numeric label names have no statistical
meaning; for reporting only, labels are ordered by the smallest observation
index in each cluster.

\paragraph{Target-selection rules.}
$\Acal_{\mathrm{app}}$ returns every unordered pair of nonempty clusters in the
partition. $\Acal_{\mathrm{2lg}}$ returns the unordered pair of the two largest
nonempty clusters, breaking size ties by the smallest observation index in the
cluster and then lexicographically. Other graph-output-based rules are allowed when specified before inference
and made deterministic. For example, a fixed external annotation may be used to
choose the pair having the largest pre-specified annotation contrast. A rule
that inspects additional raw-data summaries is not automatically covered: its
selection event must also be characterized and included.

\paragraph{Conditioning beyond the cluster pair.}
Two further data-defined events must be included in the conditioning set to make the procedure 
well-defined. They are deterministic, hence measurable, functions of the graph state; intersecting 
them with the selection event only removes additional values of $\phi$ from $S_{\Acal}$, so the 
truncated-pivot validity argument is unchanged.

\begin{assumption}[Strict spectral gap]\label{ass:gap}
We retain only $\phi$ at which $\lambda_{K+1}(L)>\lambda_K(L)$, where
$\lambda_1\le\cdots\le\lambda_n$ are the Laplacian eigenvalues, i.e.\ the $K$
eigenvalues used for the embedding are strictly separated from the remainder of
the spectrum. This excludes graphs with more than $K$ connected components and
other degeneracies where the selected bottom-$K$ invariant subspace itself is
not separated. Repeated eigenvalues inside the bottom-$K$ subspace are handled
by the canonical basis convention in
Section~\ref{sec:deterministic_pipeline}.
\end{assumption}
\begin{assumption}[Deterministic rounding]\label{ass:det}
The $k$-means rounding uses the deterministic Lloyd convention in
Section~\ref{sec:deterministic_pipeline}, including fixed initialization,
iteration order, tie-breaking, and empty-cluster handling. Within a fixed-graph
interval the embedding is identical, so the labels are identical.
\end{assumption}
Under Assumptions~\ref{ass:gap}--\ref{ass:det} and the conventions in
Section~\ref{sec:deterministic_pipeline}, the conditioning event is
$S_{\Acal}$ with invalid graph states removed. The rule $\Acal$ must match the
way the reported hypothesis was selected: the Gao--Bien--Witten
pair-appearance event uses $\Acal_{\mathrm{app}}$, whereas experiments that test
the two largest clusters use $\Acal_{\mathrm{2lg}}$.

\begin{proposition}[Validity]\label{prop:valid}
Under the model of Section~\ref{sec:setup}, under $H_0$, and for the
deterministic pipeline above, let
\[
\begin{aligned}
U&=(P_{\nuv}^{\perp}\X,\pi),\\
S_{\Acal}(U)
&=\{\phi\ge0:\ \Spec(\X'(\phi))\ \text{is valid and}\
\Cab\in\Acal(\Spec(\X'(\phi)))\},
\end{aligned}
\]
and let $E_{\Acal}=\{T(\X)\in S_{\Acal}(U)\}$ be the matched selection event.
The $p$-value \eqref{eq:pval}, evaluated over the finite-interval set
$S_{\Acal}(U)$ of Theorem~\ref{thm:finite}, satisfies
$\Prob(p\le t\mid E_{\Acal})\le t$ for all $t\in[0,1]$, with equality when
$S_{\Acal}(U)$ has positive measure almost surely on $E_{\Acal}$.
\end{proposition}

\begin{proof}[Proof sketch]
On the line \eqref{eq:path}, fixing $U=(P_{\nuv}^{\perp}\X,\pi)$ leaves only the
radius $T(\X)$ free. Under $H_0$, $\nuv^\top\X\sim
N(0,\sigma^2\|\nuv\|^2I_d)$, and Gaussian orthogonality plus isotropy give
$T(\X)\sim\sigma\|\nuv\|\chi_d$ independent of $U$. For fixed $U$, the
deterministic pipeline makes the matched selection event exactly
$\{T(\X)\in S_{\Acal}(U)\}$, so the conditional law is the same $\chi_d$ law
truncated to $S_{\Acal}(U)$. The ratio \eqref{eq:pval} is therefore the truncated
upper-tail probability and is uniform by the probability integral transform;
zero-mass truncation sets are assigned $p=1$. Full details are in the supplement.
\end{proof}

\subsection{Feature-wise selective inference}\label{sec:featurewise}
The global test asks whether two selected clusters differ in their overall
$d$-dimensional mean. In high-dimensional applications it is often more useful
to ask which coordinates explain the selected separation. For feature
$j\in\{1,\dots,d\}$, consider
\begin{equation}\label{eq:feature_null}
H_{0j}:\ \nuv^\top\bmu_{\cdot j}=0,
\qquad
Z_j:=\nuv^\top\X_{\cdot j}.
\end{equation}
Under $H_{0j}$, $Z_j\sim N(0,\sigma^2\|\nuv\|^2)$. Condition on the other
features $\X_{\cdot,-j}$ and on the nuisance projection
$P_{\nuv}^{\perp}\X_{\cdot j}$. The one-dimensional path that varies only the
tested feature is
\begin{equation}\label{eq:feature_path}
\X_j'(\phi)
\ :=\
\X+\frac{\phi-Z_j}{\|\nuv\|^2}\,\nuv\,e_j^\top,
\qquad \phi\in\R,
\end{equation}
where $e_j$ is the $j$th coordinate vector. Then
$\nuv^\top\X_j'(\phi)_{\cdot j}=\phi$, all other coordinates are unchanged, and
the nuisance $P_{\nuv}^{\perp}\X_{\cdot j}$ is preserved. Define the
feature-wise selection set
\begin{equation}\label{eq:Sfeature}
S_{\Acal,j}
\ :=\
\{\phi\in\R:\ \Spec(\X_j'(\phi))\ \text{is valid and}\
\Cab\in\Acal(\Spec(\X_j'(\phi)))\}.
\end{equation}
Because \eqref{eq:feature_path} is affine in $\phi$, every pairwise squared
distance along the path is again a quadratic polynomial in $\phi$. Hence
Theorem~\ref{thm:finite} and Corollary~\ref{cor:neighborhood} apply verbatim:
for unweighted $\varepsilon$- and $k$-NN graphs, $S_{\Acal,j}$ is an exact finite
union of intervals whose finite endpoints are among the corresponding
data-dependent quadratic roots.

\begin{proposition}[Feature-wise validity]\label{prop:feature}
Under the model of Section~\ref{sec:setup}, $H_{0j}$, and
the deterministic pipeline above, conditional on the matched selection event
and the feature-wise nuisance variables, $Z_j$ has a
$N(0,\sigma^2\|\nuv\|^2)$ distribution truncated to $S_{\Acal,j}$. Consequently
the two-sided feature-wise selective $p$-value
\begin{equation}\label{eq:feature_pval}
p_j
=
\frac{
\Prob\big(|Y_j|\ge |Z_j^{\mathrm{obs}}|,\ Y_j\in S_{\Acal,j}\big)}
{\Prob\big(Y_j\in S_{\Acal,j}\big)},
\qquad
Y_j\sim N(0,\sigma^2\|\nuv\|^2),
\end{equation}
is super-uniform under $H_{0j}$; one-sided alternatives use the corresponding
upper or lower tail of the same truncated normal law.
\end{proposition}

\begin{proof}[Proof sketch]
Fix the selected pair and condition on
$U_j=(\X_{\cdot,-j},P_{\nuv}^{\perp}\X_{\cdot j})$. Under $H_{0j}$,
$Z_j=\nuv^\top\X_{\cdot j}\sim N(0,\sigma^2\|\nuv\|^2)$ and is independent of
$U_j$ by Gaussian orthogonality and the spherical feature model. For fixed
$U_j$, the matched selection event is $\{Z_j\in S_{\Acal,j}(U_j)\}$, and
Corollary~\ref{cor:neighborhood} makes this set a finite union of intervals for
the graph classes considered here. The truncated-normal probability integral
transform gives \eqref{eq:feature_pval}; one-sided tails are analogous. Full
details are in the supplement.
\end{proof}

\begin{remark}[Multiple features]
If all features are tested for a fixed selected pair, Proposition~\ref{prop:feature}
gives valid marginal selective $p$-values for marker or driver features. The
subsequent multiplicity correction is a separate layer: Bonferroni or
Benjamini--Yekutieli control is valid under arbitrary dependence, while more
powerful false-discovery-rate procedures require assumptions on the dependence
among the selective $p$-values. If the analyst first screens features and then
tests only the screened subset, that screening rule should also be included in
the selection event or handled by sample splitting.
\end{remark}

\subsection{Direct extensions}\label{sec:extensions}
The finite-state proof is stable under deterministic operations that are fixed
on each interval where the neighborhood graph is fixed. We therefore keep the
main construction focused on the fixed-$K$ spherical case above, and record only
the application-facing extensions here. The supplement gives the full whitening
and auxiliary-coordinate statements and proofs.

\begin{remark}[Known feature covariance]\label{rem:whiten}
If the rows are independent Gaussian observations with known covariance
$\Sigma\succ0$, apply the entire pipeline to the whitened matrix
$\tilde\X=\X\Sigma^{-1/2}$: build
the graph in the Mahalanobis metric and use the Mahalanobis difference-in-means
statistic. Then $\tilde\X$ satisfies the unit-scale spherical model, so
Proposition~\ref{prop:valid} applies with reference law $\|\nuv\|\chi_d$
truncated to the finite-state selection set computed after whitening. Whitening
only the statistic while keeping a Euclidean graph on the unwhitened data is not
covered, and per-coordinate feature-wise pivots under general $\Sigma$ remain
outside the present theorem.
\end{remark}

\paragraph{Data-driven number of clusters.}
Fix a maximum $K_{\max}$ and use the deterministic eigengap rule
\begin{equation}\label{eq:eigengap}
\hat K(G)\ :=\ \operatorname*{arg\,max}_{k\in\{1,\dots,K_{\max}\}}
\big(\lambda_{k+1}(L)-\lambda_k(L)\big),
\end{equation}
with ties broken by the smallest such $k$. Because the Laplacian is fixed on
each finite-state interval, the whole spectrum and hence $\hat K$ are constant
there.
\begin{proposition}[Exact inference with a data-driven $K$]\label{prop:datak}
Augment the deterministic pipeline of Section~\ref{sec:deterministic_pipeline}
so that the embedding dimension is chosen by \eqref{eq:eigengap}, and add
$\{\hat K(\phi)=K_{\mathrm{obs}}\}$ to the selection event, where
$K_{\mathrm{obs}}=\hat K(G(\x^{\mathrm{obs}}))$. Under the model of
Section~\ref{sec:setup} and $H_0$, the augmented selection set
\[
S_{\Acal}^{K}\ :=\ \{\phi\ge0:\ \Spec(\X'(\phi))\ \text{is valid},\
\hat K(\phi)=K_{\mathrm{obs}},\ \text{and}\ \Cab\in\Acal(\Spec(\X'(\phi)))\}
\]
is a finite union of intervals whose finite endpoints are among the
data-dependent quadratic roots, and the $p$-value
\eqref{eq:pval} evaluated over $S_{\Acal}^{K}$ is super-uniform.
\end{proposition}
The proof, given in the supplement, is immediate from Lemma~\ref{lem:detinterval}
and Proposition~\ref{prop:valid}: the event
$\{\hat K(\phi)=K_{\mathrm{obs}}\}$ is just one more measurable,
path-piecewise-constant selection event. The reported experiments fix $K=3$;
\eqref{eq:eigengap} is the validity-preserving way to let the data set $K$.

\paragraph{Graphs built from auxiliary coordinates.}
If the graph is a deterministic function of auxiliary coordinates independent of
the tested features---for example, spatial coordinates used to cluster cells
before testing expression---then the graph is held fixed along the expression
perturbation path. The selection event is independent of the contrast, the
truncation set is the whole ray (or all of $\mathbb R$ for feature-wise tests),
and the ordinary test is already exact. This boundary case is stated and proved
in the supplement.

\paragraph{Algorithm.}
Algorithm~\ref{alg:main} constructs the finite-interval representation of
$S_{\Acal}$. Two optimizations reduce computation without changing the output: (a) \emph{spectator sparsity}---under
$\sigma^2 I$, $q_i\ne0$ only for $i\in C_a\cup C_b$, so only breakpoints from
candidate pairs with at least one endpoint in $C_a\cup C_b$ need be enumerated;
spectator--spectator distances are constant along the path; (b) \emph{graph
deduplication}---consecutive intervals frequently share the same graph (the
swapped pair lies deep in the ordering), so the eigendecomposition and rounding
are recomputed only when $G$ actually changes.

\begin{algorithm}[t]
\caption{Finite-state computation of the selective $p$-value}\label{alg:main}
\medskip\small
\begin{minipage}{0.98\linewidth}
\textbf{Input:} $\X$, $\sigma$, $K$, neighborhood size; observed labels and
pair $(C_a,C_b)$; deterministic target-selection rule $\Acal$.
\begin{enumerate}
\item Form $\nuv$ and the affine path coefficients $p_i,q_i$.
\item Compute all graph-change roots from
$\alpha_{ij},\beta_{ij},\gamma_{ij}$ in Eq.~\eqref{eq:quad}.
\item Set
$B\gets\{0\}\cup\{\text{positive finite roots returned by the solver}\}
\cup\{\infty\}$; deduplicate roots by floating-point equality and sort $B$.
\item Initialize $S_{\Acal}\gets\emptyset$ and
$g_{\mathrm{prev}}\gets\bot$.
\item For each consecutive interval $(\ell,u)$ in $B$:
  \begin{enumerate}
  \item Choose $\phi^\star$ as the midpoint of $(\ell,u)$, or $\ell+1$ when
  $u=\infty$.
  \item Build $G$ from $\{d_{ij}(\phi^\star)^2\}$ and set
  $g\gets$ the byte representation of the binary adjacency matrix.
  \item If $g=g_{\mathrm{prev}}$, verify matrix equality and reuse the previous decision.
  \item Otherwise, check Assumption~\ref{ass:gap}. If it fails, set
  $\mathrm{ok}\gets\mathrm{false}$; if it passes, compute the canonical
  embedding, run deterministic $k$-means, and set
  $\mathrm{ok}\gets(\Cab\in\Acal(\Cc(\X'(\phi^\star))))$.
  \item Set $g_{\mathrm{prev}}\gets g$. If $\mathrm{ok}$, add $[\ell,u]$ to
  $S_{\Acal}$.
  \end{enumerate}
\item Merge adjacent accepted intervals using the fixed boundary tolerance.
\item Return the truncated-$\chi_d$ $p$-value over $S_{\Acal}$.
\end{enumerate}
\end{minipage}
\end{algorithm}

\paragraph{Homotopy line search.}
We implement Algorithm~\ref{alg:main} as a homotopy that walks $\phi$ from $0$
to $\infty$ and recomputes the embedding and rounding \emph{only} when the graph's
binary adjacency representation changes, reusing the decision across the (often many) intervals that share a
graph. This is consequential for $k$-NN graphs, where most rank-swap breakpoints
occur deep in the distance ordering and leave the top-$k$ graph unchanged: in our
experiments the number of spectral solves was roughly $15\times$ smaller than the
number of breakpoint intervals (Section~\ref{sec:experiments}). For
$\varepsilon$-graphs every breakpoint is an edge flip, so solves equal intervals.

\paragraph{Floating-point conventions and numerical audit.}
The theorem is an exact real-arithmetic statement; the implementation is a
floating-point proof of concept. In the initially submitted solver, coefficients with
$|a|<10^{-12}$ are treated as linear, negative computed discriminants are
discarded rather than clipped, only positive finite roots above $10^{-12}$ are
retained, and candidate roots are deduplicated by exact floating-point equality.
The eigengap tolerance is $10^{-7}$, and adjacent accepted intervals are merged
when their displayed boundary difference is below $10^{-9}$. Infinite intervals
use $\ell+1$ as an interior evaluation point. Graph reuse keys on the binary
adjacency byte string and verifies matrix equality before reusing a decision.
The revised camera-ready code also makes the theoretical lexicographic $k$-NN tie rule
explicit; the audit below found no target or validity changes from that update.
Endpoint membership is immaterial for the continuous reference laws.

We audited 400 fixed datasets against six alternative numerical policies. Of
2,400 attempted policy--dataset comparisons, 2,394 had an evaluable submitted
target. No alternative changed overall evaluation status; among evaluable cases,
neither the selected target nor eigengap validity changed. In 484 comparisons
the two computed selection sets had nonzero symmetric-difference probability
mass, but every difference was confined to extremely narrow intervals adjacent
to selection-set boundaries; the largest such mass was
$5.35\times10^{-16}$, and the largest absolute $p$-value difference was
$1.10\times10^{-13}$. A 100-digit root/graph-state audit matched the submitted
state sequence in 48/48 high-risk cases. These checks are empirical sensitivity
evidence, not interval-arithmetic certification: eigenspaces and rounding remain
float64, so the implementation does not provide certified upper/lower
truncation probabilities. Full policies and results are in the supplement.

\paragraph{Conditioning level: over-conditioning vs.\ parametric programming.}
A single pass can yield valid $p$-values at several conditioning levels. The
\emph{over-conditioned} test fixes the realized graph,
$S_{\mathrm{oc}}=\{\phi: G(\phi)=G(\x^{\mathrm{obs}})\}$; the
\emph{matched-rule} test uses $S_{\Acal}$ for the target-selection rule actually
used to report $\Cab$; and the pair-appearance special case is
$S_{\mathrm{app}}=\{\phi:\Cab\in\Acal_{\mathrm{app}}(\Cc(\X'(\phi)))\}$.
Since the graph determines the clustering and the target-selection rule is a
deterministic function of the clustering, all three sets are produced by the
same homotopy. The less one conditions, the more power is available, but validity
requires that the conditioning event match the reported hypothesis: if the
protocol tests the two largest clusters, the correct power-preserving set is
$S_{\mathrm{2lg}}$, not merely $S_{\mathrm{app}}$.

\section{Experiments}\label{sec:experiments}
\paragraph{Analysis protocol.}
Unless a robustness cell explicitly changes one component, graph type,
neighborhood size, Laplacian, $K$, rounding seed, target rule, preprocessing,
and tested features are fixed before inspecting the tested outcome. The primary
and power experiments use OR-symmetrized $k$-NN with $k=5$,
$L_{\mathrm{sym}}$, $K=3$, deterministic rounding, and the two-largest rule.
The $\varepsilon$ robustness threshold is fixed at
$1.3\sqrt{2d\sigma^2}$. Other data-dependent tuning choices must be fixed in
advance, selected on independent data, or incorporated into the selection event.

\subsection{Primary null calibration}
\label{sec:primary-calibration}
We first verify type-I control under the \emph{global null} (all $\bmu_i$ equal,
so no true clusters), where valid selective $p$-values must be uniform on
$[0,1]$. The primary configuration uses the full Algorithm~\ref{alg:main}:
OR-symmetrized $k$-NN graph with $k=5$, $L_{\mathrm{sym}}$, $K=3$, deterministic
$k$-means rounding, and the matched two-largest-cluster target rule
$\Acal_{\mathrm{2lg}}$. The primary setting is $n=32,d=5$; we retain
$n=18,d=5$ as a small supporting configuration, with $5000$ usable null
replicates in each setting.
A replicate is ``usable'' if the deterministic pipeline returns the specified
target and passes the validity checks; failures are reported separately.

\begin{table}[t]
\centering
\caption{Primary type-I calibration under the global null for $k$-NN,
$L_{\mathrm{sym}}$, $K=3$, and matched $\Acal_{\mathrm{2lg}}$ conditioning
($d=5$, $\sigma=1$). Brackets give exact binomial 95\% confidence intervals for
tail probabilities.}
\label{tab:primary-calibration}
\small
\resizebox{\textwidth}{!}{\input{si_spectral_primary_calibration_table.tex}}
\end{table}

\begin{figure}[t]
\centering
\includegraphics[width=0.50\textwidth]{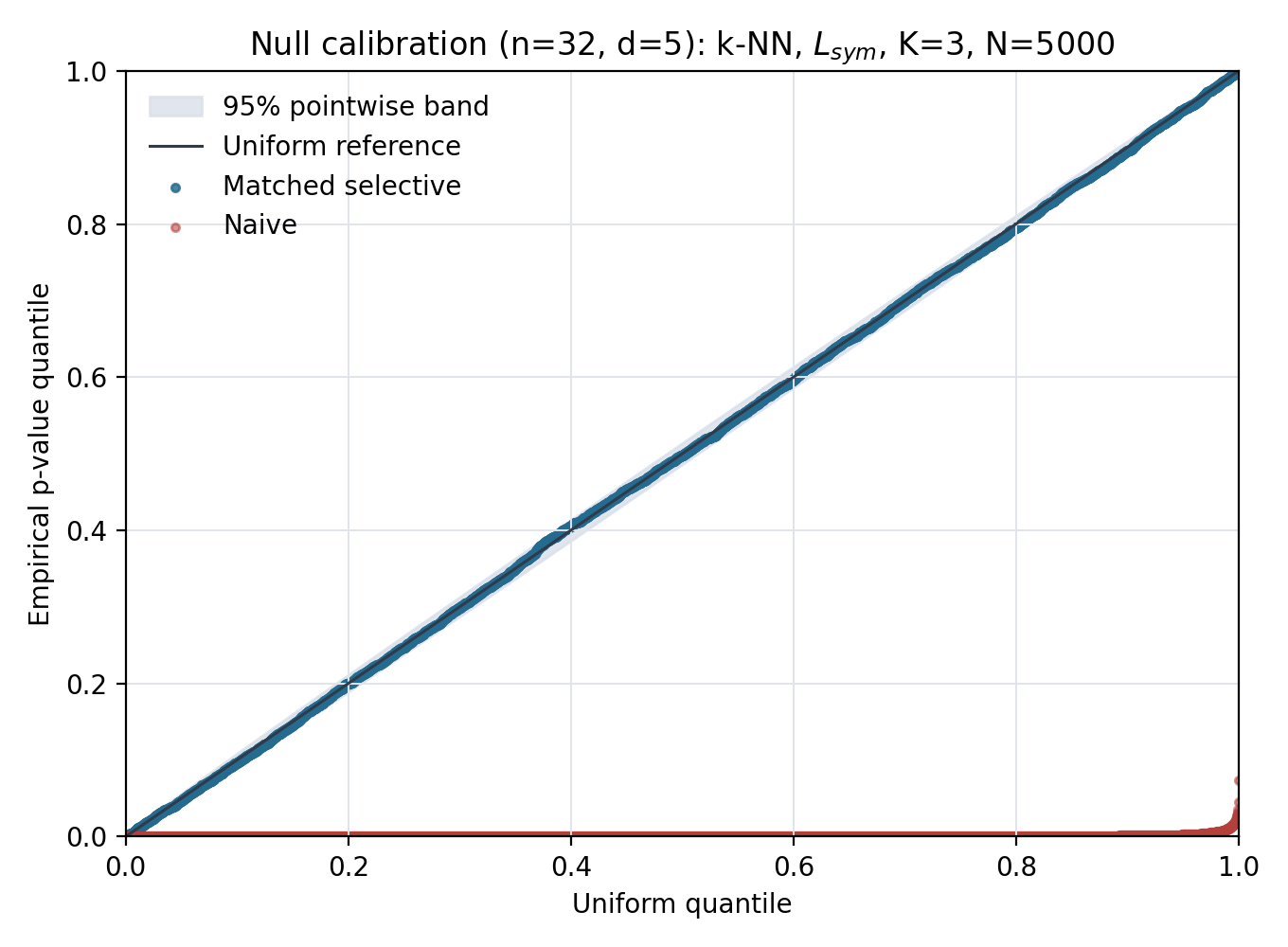}
\caption{Quantile--quantile plot under the primary $n=32,d=5$ global-null
configuration ($5000$ attempted and usable replicates). The gray band is the
pointwise 95\% beta band for uniform order statistics. Matched selective
$p$-values track the diagonal, whereas the naive selected-cluster test collapses
to the floor. The supporting $n=18$ plot is in the supplement.}
\label{fig:validation}
\end{figure}

The selective $p$-values are indistinguishable from uniform: KS $p$-values are
$0.911$ for $n=32$ and $0.222$ for $n=18$, and the empirical rejection rates at
levels $.01$, $.05$, and $.10$ all fall within their binomial confidence
intervals. No observed replicate failed because of the spectral gap,
zero-degree vertices, empty-cluster output, or numerical truncation-mass issues.
The naive test rejected in $99.98\%$ of the primary $n=32$ null replicates and
$95.2\%$ of the supporting $n=18$ replicates. For the primary $n=32$ setting,
the average number of candidate intervals was $7317$, but graph deduplication
reduced this to $119$ spectral
solves; the supporting $n=18$ setting had $1277$ intervals and $69$ solves.
Thus the number of interval blocks is not the number of eigendecompositions,
so candidate intervals and eigendecompositions should be distinguished,
although the explicit interval walk remains costly.

\subsection{Laplacian variants, graph types, and conditioning level}
\label{sec:variants}
We next exercise the unified procedure across graph type ($\varepsilon$,
$k$-NN), Laplacian ($L_{\mathrm{sym}}$, $L_{\mathrm{rw}}$), and conditioning
level under the global null. The supplement reports $2000$ usable replicates per
cell for both $n=16$ and $n=32$ ($d=5$, $K=3$). The matched rows use the same
two-largest target-selection rule as the primary calibration; the
graph-conditioned rows fix the realized graph for the main $k$-NN,
$L_{\mathrm{sym}}$ configuration. All level $.05$ conditional tail-rate confidence intervals cover the
nominal level. Under the target-usable stopping rule, the $n=16$
$\varepsilon$ random-walk and symmetric cells required 2,099 and 2,101 attempts
to obtain 2,000 usable replicates: usability was $0.953$ and $0.952$, conditional
rejection was $0.050$ and $0.0555$, and joint usable-and-reject probability was
$0.0476$ and $0.0528$. The extra attempts were observed-gap failures. All
$k$-NN cells and all $n=32$ cells were usable in 2,000/2,000 attempts. Because
these runs stopped at 2,000 usable datasets, the attempt-based rates are
diagnostics rather than pre-specified fixed-attempt binomial estimates.

\begin{figure}[t]
\centering
\includegraphics[width=\textwidth]{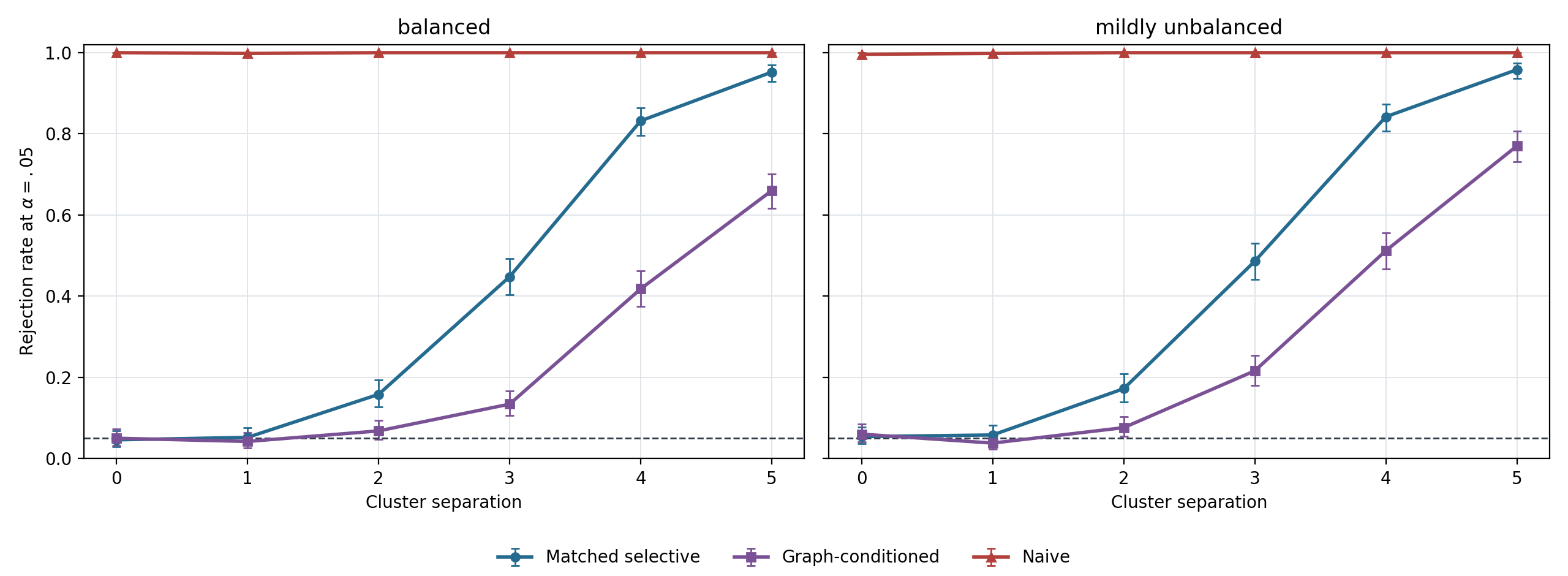}
\caption{Power curves for three Gaussian clusters under $k$-NN,
$L_{\mathrm{sym}}$, $K=3$, and the two-largest target rule, with $500$ fixed attempts per separation (all usable) and exact binomial
95\% confidence intervals. Matched
target-selection conditioning retains substantially more power than fixing the
realized graph, while the naive test is already invalid at $\delta=0$.}
\label{fig:power-curves}
\end{figure}

\paragraph{Power: conditioning level matters.}
We next vary signal strength in three-cluster Gaussian alternatives
($k$-NN, $L_{\mathrm{sym}}$, $K=3$, $d=5$). The cluster centers are
$0$, $\delta e_1$, and $\delta e_2$, with
$\delta\in\{0,1,2,3,4,5\}$. We use $500$ fixed attempts per separation for a
balanced design with cluster sizes $(10,10,10)$ and a mildly unbalanced design
with sizes $(6,10,14)$. Figure~\ref{fig:power-curves} compares the naive test,
realized-graph conditioning, and matched $\Acal_{\mathrm{2lg}}$ conditioning.
All 500/500 attempts were usable in every cell, so conditional rejection
and operational joint discovery coincide in these runs; the finite-sample lower
95\% confidence bound on usability in each cell is $0.993$. The plotted
uncertainty bars are exact binomial 95\% confidence intervals for rejection.
At $\delta=0$, the selective tests are near nominal while the naive test rejects
almost always because the clusters were selected from the same noise. As
separation grows, matched conditioning is materially more powerful than fixing
the realized graph: at $\delta=4$, matched power is $0.832$ (balanced) and
$0.842$ (unbalanced), versus $0.418$ and $0.512$ for graph conditioning; at
$\delta=5$ the matched powers are $0.952$ and $0.958$.
The supplement also reports a valid Gaussian data-thinning baseline, but it
targets clusters selected from an independent noisy copy $X^{\mathrm{sel}}$
rather than the spectral clustering actually returned on $\X$. In some power
cells this different target is easier; for example, in Supplementary Table~\ref{tab:supp-thinning} the
balanced $\delta=3$ thinning rejection rate is $0.728$, compared with $0.448$
for the matched selective test.

\paragraph{Feature-wise partial null in structured data.}
We generated $n=30,d=5$ spherical Gaussian data with three balanced groups,
signal in the first two coordinates, and a pre-specified fifth feature having
mean zero in every true group. For separations $\delta=0,2,4,5$, all
1,000/1,000 attempts were usable. Selective level-.05 rejection was
$0.050,0.048,0.043,0.054$, respectively, versus naive rates
$0.447,0.275,0.059,0.051$. Median truncation masses were
$0.0301,0.101,0.917,1.000$. Thus calibration persisted both when the selection
set was restrictive and when genuine cluster structure was present; full exact
confidence intervals and truncation-mass quantiles are in the supplement.

\paragraph{Model-scope diagnostics.}
In a fixed-attempt grid of 500 datasets per setting, rejection was $0.052$
under the theorem-backed spherical Gaussian model. Outside the theorem, row
heteroskedasticity also gave $0.052$; row correlation $\rho=0.2$ gave a
conservative $.05$ tail rate of $0.028$ but a non-uniform full distribution.
For feature correlations $\rho=0.1,0.25,0.5$, the rates were $0.066$, $0.068$,
and $0.144$. The submitted standardized-$t_5$ stress test rejected at $0.013$
in 300 datasets. These rows are diagnostics and not theorem extensions; feature
correlation can be materially anti-conservative. Pilot direct evaluations found
no selection-set mismatch at 746 checked points in the new partial-null runs.

\subsection{Computational scaling}
\label{sec:scaling}
Let $M$ denote candidate intervals and $U$ the distinct graph states that
require spectral solves. Coefficient construction costs $O(n^2d)$ time and
$O(n^2)$ memory; the explicit root list has $M=O(n^2)$ for $\varepsilon$
graphs and $M=O(n^3)$ for $k$-NN, with $O(M\log M)$ sorting and
$O(n^2+M)$ storage. The current code rebuilds each graph, costing
$O(Mn^2)$ for $\varepsilon$ and $O(Mn^2\log n)$ for $k$-NN, then performs
$U$ dense $O(n^3)$ spectral solves; $U\le M$. Testing $F$ features repeats the
feature-wise homotopy approximately $F$ times.

A component diagnostic (three fixed attempts per graph/$n$ cell) confirms the
current bottleneck. For $k$-NN at $n=32,64,128$, mean candidate intervals were
6,890, 59,083, and 497,839, versus 107, 240, and 482 spectral solves; mean times
were 0.18, 5.89, and 244.59 seconds, with the graph walk accounting for
83.9\%, 97.3\%, and 99.6\%. For $\varepsilon$ graphs, the corresponding
$n=32,64,128$ means were 273, 1,186, and 4,989 intervals/solves and 0.06, 0.56,
and 6.54 seconds, with eigendecomposition/rounding accounting for
87.8--94.5\%. Note that these measurements characterize the recorded proof-of-concept
engine, not hardware-independent constants or an optimized-complexity claim.
Larger problems require skipping irrelevant rank swaps, incremental graph
maintenance, and warm-started or partial eigensolvers.

\subsection{Working-model PBMC algorithmic illustration}
\label{sec:pbmc}
We use the PBMC 3k dataset~\citep{wolf2018scanpy} to illustrate execution of the
feature-wise homotopy on a fixed preprocessed matrix; this is not to support
confirmatory biological inference. The protocol fixes a pre-specified 16-gene
panel, log-normalized library-size-adjusted counts, marginal gene
standardization, a fixed-seed subsample of 60 cells, $k=5$, $K=3$, and the
matched two-largest rule. The observed cluster sizes are $(31,21,8)$, and the 16
homotopies finish in 55.5 seconds.

The calculation sets the working scale to $\sigma=1$ after standardization;
this is not a known sampling variance, and marginal standardization does not
remove gene--gene correlation. The misspecification grid above shows that
feature correlation can be strongly anti-conservative. Accordingly, the
displayed quantities are working-model selective values and not valid confirmatory
scientific $p$-values under a defensible PBMC sampling model. We make no marker
significance or biological conclusion. The fixed-panel table and diagnostic
plots are moved to the supplement, where they are labeled with the same caveat.


\section{Discussion}\label{sec:discussion}
\paragraph{Conditioning and power.}
The same graph-state walk yields several valid conditioning levels: fixing the
realized graph is cheapest but most conservative, while conditioning only on the
matched target-selection rule $\Cab\in\Acal(C(\cdot))$ preserves the reported
hypothesis with more power. Figure~\ref{fig:power-curves} illustrates the gap: at
separation $\delta=4$, matched-rule conditioning gains 33--41 percentage points
over graph conditioning without extra graph enumeration. This is the same
conditioning/power trade-off emphasized in parametric-programming selective
inference and related post-selection problems.

\paragraph{Different targets and pipeline use.}
Data thinning is a valid baseline, but it clusters a randomized copy
$X^{\mathrm{sel}}$ and tests on an independent copy $X^{\mathrm{inf}}$; our
selective test instead makes inference for the deterministic spectral clustering
returned on the observed matrix $X$. Theorem~\ref{thm:finite} and
Corollary~\ref{cor:neighborhood} also make the unweighted-graph spectral step a
natural component for multi-stage clustering-pipeline selective inference, where
linear/quadratic selection events are already handled for steps such as feature
selection, $k$-means, and DBSCAN.

\paragraph{Assumptions, numerical status, and limitations.}
The finite-state graph reduction does not require known $\sigma$, but the stated
truncated-$\chi$ and truncated-normal reference laws do. An independent or
same-data scale estimate cannot simply be plugged in and called exact; an exact
studentized construction would require a new conditional-law derivation. When a
known covariance $\Sigma$ is available, validity applies only if the analysis
itself is defined by whitening first and clustering/testing in the resulting
Mahalanobis metric; it does not validate the Euclidean clusters that would have
been obtained before whitening, and per-coordinate pivots under general or
unknown covariance remain open. Fixed preprocessing and deterministic rounding
are required. Data-driven $K$ under Proposition~\ref{prop:datak} is covered,
but feature selection, graph type, neighborhood size, normalization, and other
tuning choices must be fixed, independently selected, or added to the selection
event. Smooth kernels break the finite quadratic-root representation. Finally,
the numerical audit supports stability on the audited datasets but is not
end-to-end certification, and the current implementation is a proof of concept
for the studied problem sizes rather than a practical large-scale system.

\section{Conclusion}
Exact selective inference for spectral clustering is tractable at the
statistical-theorem level, and the eigendecomposition is not the obstruction.
For unweighted neighborhood graphs, clustering is piecewise constant along the
selective perturbation, with finite breakpoints among roots of data-dependent
quadratics. This yields finite-sample exact truncated pivots in real arithmetic
under the stated known-scale spherical Gaussian model. The same graph-state
argument gives feature-wise tests and supplies the previously missing spectral
component for clustering-pipeline selective inference. Smooth kernels,
unknown/correlated covariance, adaptive preprocessing, scalable implementation,
and end-to-end numerical certification remain outside the present guarantee.

\footnotesize
\setlength{\bibsep}{0pt plus 0.3ex}

\clearpage
\normalsize
\setcounter{section}{0}
\setcounter{subsection}{0}
\setcounter{subsubsection}{0}
\setcounter{figure}{0}
\setcounter{table}{0}
\setcounter{equation}{0}
\setcounter{theorem}{0}
\setcounter{suppprop}{0}
\renewcommand{\thesection}{S\arabic{section}}
\renewcommand{\thesubsection}{S\arabic{section}.\arabic{subsection}}
\renewcommand{\thesubsubsection}{S\arabic{section}.\arabic{subsection}.\arabic{subsubsection}}
\renewcommand{\thefigure}{S\arabic{figure}}
\renewcommand{\thetable}{S\arabic{table}}
\renewcommand{\theequation}{S\arabic{equation}}
\renewcommand{\thetheorem}{S\arabic{theorem}}
\renewcommand{\thesuppprop}{S\arabic{suppprop}}
\makeatletter
\@ifundefined{theHsection}{}{\renewcommand{\theHsection}{supp.\arabic{section}}}
\@ifundefined{theHsubsection}{}{\renewcommand{\theHsubsection}{supp.\arabic{section}.\arabic{subsection}}}
\@ifundefined{theHsubsubsection}{}{\renewcommand{\theHsubsubsection}{supp.\arabic{section}.\arabic{subsection}.\arabic{subsubsection}}}
\@ifundefined{theHfigure}{}{\renewcommand{\theHfigure}{supp.\arabic{figure}}}
\@ifundefined{theHtable}{}{\renewcommand{\theHtable}{supp.\arabic{table}}}
\@ifundefined{theHequation}{}{\renewcommand{\theHequation}{supp.\arabic{equation}}}
\@ifundefined{theHtheorem}{}{\renewcommand{\theHtheorem}{supp.\arabic{theorem}}}
\@ifundefined{theHsuppprop}{}{\renewcommand{\theHsuppprop}{supp.\arabic{suppprop}}}
\makeatother

\phantomsection
\pdfbookmark[0]{Supplementary Material}{supplementary-material}
\begin{center}
{\Large\bfseries Supplementary Material\par}
\vspace{0.6em}
{\large Exact Selective Inference for Spectral Clustering via\par
Piecewise-Constant Graph Perturbation\par}
\vspace{0.7em}
{Naotoshi Nakamura\par}
\end{center}
\vspace{1.2em}

\section{Expanded Proofs and Technical Details}

\subsection{Proof of Theorem~\ref{thm:finite} and Corollary~\ref{cor:neighborhood}}

\begin{proof}[Proof of Lemma~\ref{lem:detinterval}]
On an interval where $G(\phi)=G_0$, every downstream operation is evaluated at
the same graph state. The Laplacian construction, gap/validity check,
canonical embedding map, rounding rule, label-set representation, and
target-selection rule are fixed measurable functions of that state. Their
composition is therefore a single constant value on the interval.
\end{proof}

\begin{proof}[Proof of Theorem~\ref{thm:finite}]
Let
\[
B=\{\phi\ge0:h_m(\phi)=0\ \text{for at least one nonzero } h_m\}.
\]
Because there are finitely many nonzero polynomials and each has finite degree,
$B$ is finite. On any connected component $I$ of $[0,\infty)\setminus B$, every
sign $\operatorname{sign}h_m(\phi)$ is constant, so the graph selector returns a
fixed graph $G_I$ throughout $I$. By the deterministic spectral pipeline and the
strict-gap convention, the graph-to-label map either returns a single partition
or the invalid symbol. Lemma~\ref{lem:detinterval} therefore makes the target
decision $\mathds{1}\{\Cab\in\Acal(\Cc(\X'(\phi)))\}$ constant on $I$, with
invalid graph states contributing decision zero. Hence $S_{\Acal}$ is the union
of those components for which the constant decision is one, up to finitely many
endpoints in $B$. These endpoints have no effect on the truncated $\chi$ or
truncated normal pivots because the reference laws are continuous.

If the coefficients of the $h_m$'s are algebraic, each finite endpoint is a
root of a polynomial with algebraic coefficients. Thus every finite endpoint is
an algebraic number.
\end{proof}

\begin{proof}[Proof of Corollary~\ref{cor:neighborhood}]
Along the selective path \eqref{eq:path}, row $i$ can be written as
\[
\x_i'(\phi)=p_i+\phi q_i,\qquad
p_i=\x_i-\frac{\nu_i}{\|\nuv\|^2}\nuv^\top\X,\qquad
q_i=\frac{\nu_i}{\|\nuv\|^2}\pi .
\]
For $i\ne j$,
\[
d_{ij}(\phi)^2
=\|\x_i'(\phi)-\x_j'(\phi)\|^2
=\alpha_{ij}\phi^2+\beta_{ij}\phi+\gamma_{ij},
\]
where
\[
\alpha_{ij}=\|q_i-q_j\|^2,\qquad
\beta_{ij}=2(p_i-p_j)^\top(q_i-q_j),\qquad
\gamma_{ij}=\|p_i-p_j\|^2.
\]
Thus every pairwise squared distance is a polynomial of degree at most two.

For an $\varepsilon$-graph, the edge $\{i,j\}$ is determined by the predicate
$d_{ij}(\phi)^2\le\varepsilon^2$, whose boundary is the quadratic equation
$d_{ij}(\phi)^2-\varepsilon^2=0$. Hence the graph can change only at roots of
these $O(n^2)$ quadratic equations.

For an OR-symmetrized $k$-NN graph, node $i$ orders candidates by
$(d_{ij}(\phi)^2,j)$, with the index tie-breaker fixed in advance. Away from
ties, the top-$k$ set is determined by the order predicates
$d_{ij}(\phi)^2<d_{im}(\phi)^2$ for $j,m\ne i$. Each boundary is
\[
d_{ij}(\phi)^2-d_{im}(\phi)^2=0,
\]
again a polynomial equation of degree at most two. There are
$O(n^3)$ such candidate order-swap equations. Between consecutive candidate
roots all edge predicates, order predicates, and hence the graph are constant.
Theorem~\ref{thm:finite} then gives the finite union of intervals. The finite
endpoints are roots of degree-$\le2$ polynomials with coefficients determined by
the data.
\end{proof}

\begin{remark}[Degenerate quadratic coefficients]
If $\alpha_{ij}=0$, the corresponding distance or order boundary is linear or
constant. Constant nonzero predicates contribute no breakpoint. Identically zero
boundaries correspond to exact distance ties on the path; the deterministic
endpoint and lexicographic tie conventions assign these cases to a fixed graph
state, and the boundary itself is null for the continuous pivot laws.
\end{remark}

\subsection{Proof of Proposition~\ref{prop:valid}}

\begin{proof}
Let $T(\X)=\|\nuv^\top\X\|$ and $\pi=\nuv^\top\X/T(\X)$ on $\{T>0\}$, and let
\[
P_{\nuv}^{\perp}=I_n-\frac{\nuv\nuv^\top}{\|\nuv\|^2},
\qquad
U=(P_{\nuv}^{\perp}\X,\pi).
\]
The data decompose as
\[
\X
=P_{\nuv}^{\perp}\X+\frac{\nuv(\nuv^\top\X)^\top}{\|\nuv\|^2}
=P_{\nuv}^{\perp}\X+\frac{\nuv(T(\X)\pi)^\top}{\|\nuv\|^2}.
\]
Conversely, replacing $T(\X)$ by any $\phi\ge0$ gives exactly the path
\[
\X'(\phi)=P_{\nuv}^{\perp}\X+\frac{\nuv(\phi\pi)^\top}{\|\nuv\|^2},
\]
so $(U,\phi)$ reconstructs the data matrix on the selective line.

Under the null hypothesis,
\[
Z:=\nuv^\top\X\sim N(0,\sigma^2\|\nuv\|^2 I_d).
\]
For each feature, $Z$ is uncorrelated with $P_{\nuv}^{\perp}\X$, and joint
Gaussianity implies independence. Since $Z$ is isotropic under $H_0$, its radius
and direction are independent, with
\[
T(\X)=\|Z\|\sim \sigma\|\nuv\|\chi_d
\]
and $T(\X)$ independent of $U=(P_{\nuv}^{\perp}\X,\pi)$.

By construction of the deterministic pipeline, for fixed $U$ the matched
selection event is exactly
\[
E_{\Acal}=\{T(\X)\in S_{\Acal}(U)\},
\]
where invalid graph states are excluded from $S_{\Acal}(U)$. Therefore,
conditional on $U$ and on selection, $T(\X)$ follows a
$\sigma\|\nuv\|\chi_d$ law truncated to $S_{\Acal}(U)$. The $p$-value
\eqref{eq:pval} is the upper-tail probability under that truncated law. By the
probability integral transform it is uniform conditional on $U$ and
$E_{\Acal}$ whenever the truncation set has positive measure. If the truncation
mass is zero we set the $p$-value to one, which preserves super-uniformity.
Averaging over $U$ conditional on selection yields
\[
\Prob(p\le t\mid E_{\Acal})\le t,\qquad t\in[0,1].
\]
\end{proof}

\subsection{Proof of Proposition~\ref{prop:feature}}

\begin{proof}
Fix the selected pair $\Cab$ and its contrast vector $\nuv$. For feature $j$,
write
\[
Z_j=\nuv^\top\X_{\cdot j},\qquad s_j=\sigma\|\nuv\|,
\qquad
U_j=(\X_{\cdot,-j},P_{\nuv}^{\perp}\X_{\cdot j}).
\]
The tested column has the orthogonal decomposition
\[
\X_{\cdot j}
=P_{\nuv}^{\perp}\X_{\cdot j}
+\frac{\nuv Z_j}{\|\nuv\|^2}.
\]
Replacing $Z_j$ by a candidate scalar $\phi\in\R$ while holding $U_j$ fixed
gives the feature-wise path \eqref{eq:feature_path}. Thus $(U_j,\phi)$ is the
one-dimensional reparametrization used to evaluate the selection event for
feature $j$.

Under $H_{0j}$,
\[
Z_j\sim N(0,s_j^2).
\]
Moreover,
\[
\operatorname{Cov}(P_{\nuv}^{\perp}\X_{\cdot j},Z_j)
=\sigma^2P_{\nuv}^{\perp}\nuv=0,
\]
so Gaussianity implies $Z_j\perp P_{\nuv}^{\perp}\X_{\cdot j}$. The spherical
model also makes feature $j$ independent of $\X_{\cdot,-j}$, hence
$Z_j\perp U_j$.

For fixed $U_j$, define
\[
S_{\Acal,j}(U_j)
=\{\phi\in\R:\Spec(\X_j'(\phi))\ \text{is valid and}\
\Cab\in\Acal(\Spec(\X_j'(\phi)))\}.
\]
Because $\X=\X_j'(Z_j)$, the matched feature-wise selection event is
$\{Z_j\in S_{\Acal,j}(U_j)\}$. The finite-state graph argument applies to this
path as well because every pairwise squared distance remains quadratic in
$\phi$; hence $S_{\Acal,j}$ is a finite union of intervals for the graph classes
covered by Corollary~\ref{cor:neighborhood}.

Conditional on $U_j$ and on selection, $Z_j$ therefore follows
$N(0,s_j^2)$ truncated to $S_{\Acal,j}(U_j)$. For the two-sided statistic,
the null distribution of $|Z_j|$ induced by this truncated normal law gives the
selective tail probability \eqref{eq:feature_pval}. Applying the probability
integral transform to that induced distribution gives a uniform conditional
pivot whenever the truncation mass is positive; zero-mass cases are assigned
$p_j=1$. Averaging over the selected nuisance distribution proves
super-uniformity. One-sided pivots follow by replacing the two-sided absolute
tail with the corresponding upper or lower truncated-normal tail.
\end{proof}

\subsection{Direct extension statements and proofs}\label{supp:extensions}

\begin{suppprop}[Known covariance whitening]\label{supp:prop:whiten}
Suppose the rows are independent with $\x_i\sim N(\bmu_i,\Sigma)$ and known
$\Sigma\succ0$, and let $\tilde\X=\X\Sigma^{-1/2}$. Then the whitened rows obey
the unit-scale spherical model, and
$\bar\bmu_{C_a}=\bar\bmu_{C_b}$ if and only if
$\bar{\tilde\bmu}_{C_a}=\bar{\tilde\bmu}_{C_b}$. Consequently, applying the
pipeline to $\tilde\X$---building the graph from Mahalanobis distances and using
$T(\tilde\X)=\|\nuv^\top\X\|_{\Sigma^{-1}}$---yields, under $H_0$, an exact
selective $p$-value with reference law $\|\nuv\|\chi_d$ truncated to the
finite-state selection set computed in the whitened metric.
\end{suppprop}

\begin{proof}
Whitening is a fixed invertible linear map on features. If $\x_i\sim
N(\bmu_i,\Sigma)$ independently, then $\tilde\x_i=\Sigma^{-1/2}\x_i\sim
N(\Sigma^{-1/2}\bmu_i,\,\Sigma^{-1/2}\Sigma\,\Sigma^{-1/2})=N(\Sigma^{-1/2}\bmu_i,I_d)$,
so the rows of $\tilde\X=\X\Sigma^{-1/2}$ obey the unit-scale spherical model of
the main text. Since $\Sigma^{-1/2}$ is invertible,
$\bar\bmu_{C_a}-\bar\bmu_{C_b}=0$ iff
$(\bar\bmu_{C_a}-\bar\bmu_{C_b})\Sigma^{-1/2}=\bar{\tilde\bmu}_{C_a}-\bar{\tilde\bmu}_{C_b}=0$,
so $H_0$ is unchanged. Running the pipeline on $\tilde\X$ means the graph is
built from $\|\tilde\x_i-\tilde\x_j\|^2=(\x_i-\x_j)^\top\Sigma^{-1}(\x_i-\x_j)$
and the statistic is $T(\tilde\X)=\|\nuv^\top\tilde\X\|_2=\|\nuv^\top\X\|_{\Sigma^{-1}}$.
The selective path for $\tilde\X$,
$\tilde\X'(\phi)=\tilde\X+\tfrac{\phi-T(\tilde\X)}{\|\nuv\|^2}\nuv\tilde\pi^\top$
with $\tilde\pi=\nuv^\top\tilde\X/T(\tilde\X)$, is affine in $\phi$, so every
pairwise squared distance is quadratic in $\phi$ exactly as in
\eqref{eq:quad}. Corollary~\ref{cor:neighborhood} therefore holds verbatim for
$\tilde\X$, and Proposition~\ref{prop:valid} applied to $\tilde\X$ with
$\sigma=1$ gives the exact selective $p$-value with reference law
$\|\nuv\|\chi_d$ truncated to $S_{\Acal}$.
\end{proof}

\begin{remark}[Why graph whitening is required]
Proposition~\ref{prop:valid} requires the truncation set and the test statistic
to be measurable functions of the same one-dimensional reduction $(U,\phi)$.
Whitening only the statistic, while building a Euclidean graph on the unwhitened
$\X$, breaks this: the graph breakpoints are then quadratics in the Euclidean
distances of $\X$, whereas the reference law lives in the Mahalanobis metric, and
the two no longer share the affine line. Clustering the whitened data restores a
single metric and hence the exact reduction.
\end{remark}

\begin{proof}[Proof of Proposition~\ref{prop:datak}]
On any connected component $I$ of $[0,\infty)\setminus B$ (with $B$ the finite
breakpoint set of Theorem~\ref{thm:finite}) the graph, hence the Laplacian $L$,
is a fixed matrix; therefore its ordered spectrum
$\lambda_1(L)\le\cdots\le\lambda_n(L)$ is a fixed tuple and the eigengap
maximiser $\hat K$ of \eqref{eq:eigengap}, with its fixed tie-break, is a
constant on $I$. Augment the target decision of Lemma~\ref{lem:detinterval} by
the factor $\mathds{1}\{\hat K(\phi)=K_{\mathrm{obs}}\}$; being a deterministic
function of the frozen spectrum, it is constant on $I$, so the augmented decision
is constant on $I$. Theorem~\ref{thm:finite} then makes $S_{\Acal}^{K}$ a finite
union of intervals whose finite endpoints lie among roots of the data-dependent
polynomial predicates. Validity is
Proposition~\ref{prop:valid} verbatim: the event
$\{\hat K(\phi)=K_{\mathrm{obs}}\}$ is one more measurable selection event that is
piecewise constant along the path, so intersecting it into $E_{\Acal}$ leaves
$T(\X)\sim\sigma\|\nuv\|\chi_d$ conditionally on $U$ truncated to
$S_{\Acal}^{K}(U)$, and the probability integral transform gives super-uniformity
(with $p=1$ on zero truncation mass). The strict-gap requirement
$\lambda_{\hat K+1}>\lambda_{\hat K}$ holds automatically on any interval where
the maximal gap in \eqref{eq:eigengap} is positive.
\end{proof}

\begin{suppprop}[Auxiliary-coordinate graphs]\label{supp:prop:indepgraph}
Suppose the affinity graph is a deterministic function of auxiliary coordinates
that are held fixed along the perturbation path, or equivalently that the graph
metric ignores the perturbed directions. Then the graph is constant along the
path. If the auxiliary coordinates are independent of the tested feature matrix
$\X$, the selected contrast is independent of the test statistic, the truncation
set is vacuous, and the ordinary global and feature-wise tests are exactly valid.
\end{suppprop}

\begin{proof}
If the graph metric ignores the perturbed directions then, in the notation of
Corollary~\ref{cor:neighborhood}, every distance or order predicate entering the
selector has $q_i-q_j=0$, so its defining quadratic \eqref{eq:quad} is the
constant $\gamma_{ij}$ and contributes no breakpoint. Hence $G(\phi)\equiv G$ and
the finite-state selector has no roots on $[0,\infty)$: the single component
$[0,\infty)$ carries one graph state. If that state is valid and
$\Cab\in\Acal(\Spec(G))$ then $S_{\Acal}=[0,\infty)$, so the truncation in
\eqref{eq:pval} is vacuous and the pivot is the untruncated $\sigma\|\nuv\|\chi_d$
tail, i.e.\ the ordinary test; the identical argument on the feature-wise path
\eqref{eq:feature_path} gives $S_{\Acal,j}=\R$ and the ordinary
$N(0,\sigma^2\|\nuv\|^2)$ test. When the auxiliary coordinates defining the graph
are independent of $\X$, the induced $\nuv$ (a function of those coordinates) is
independent of $\nuv^\top\X$, so $T(\X)\sim\sigma\|\nuv\|\chi_d$ holds
conditionally on the selection and the naive test is exactly valid.
\end{proof}

\section{Primary Null Calibration Diagnostics}

\begin{figure}[H]
\centering
\includegraphics[width=0.70\textwidth]{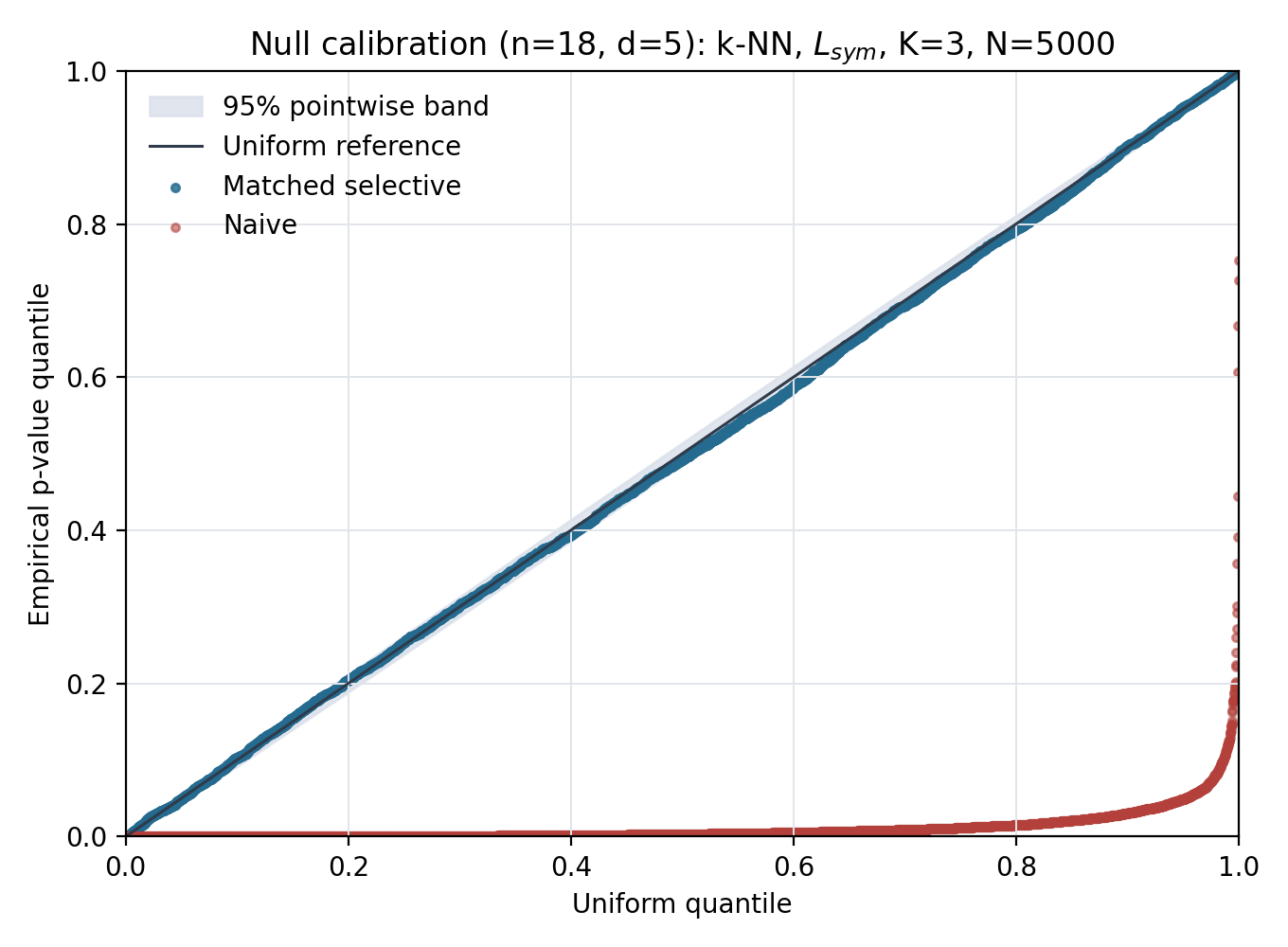}
\caption{Global-null $p$-value diagnostics for the $n=18,d=5$ primary design.
The selective curves track the uniform reference; the naive curve reflects
selection-induced anti-conservatism.}
\label{fig:supp-primary-n18}
\end{figure}

Figure~\ref{fig:supp-primary-n18} shows the smaller primary global-null design
that was displayed in the submitted main paper and is moved here to make room
for the camera-ready additions. The experiment used $5{,}000$ fixed attempts;
all $5{,}000$ were usable. The matched and graph-conditioned selective
$p$-values are close to uniform, whereas the ordinary post-clustering test is
severely anti-conservative.

\section{Floating-Point Numerical-Sensitivity Audit}
\label{supp:numerical-audit}

The finite-state characterization and pivots in the main paper are exact in
real arithmetic under the stated model. The implementation is a floating-point
proof of concept and is not an interval-arithmetic certification. The originally
submitted numerical policy treats $|a|<10^{-12}$ as linear,
discards negative computed discriminants, retains positive finite roots above
$10^{-12}$, deduplicates candidate roots by exact floating-point equality, uses
an eigengap tolerance of $10^{-7}$, and merges adjacent accepted intervals at a
boundary tolerance of $10^{-9}$.

We audited 400 fixed datasets, 50 in each of eight strata: $k$-NN global-null
settings at $n=18$ and $32$, $\varepsilon$-graph global-null settings at $n=16$
and $32$, and balanced or mildly unbalanced power settings at separations 3 and
5. Each dataset was compared with six alternative policies: a stable quadratic
formula, long-double root arithmetic, relaxed root handling, eigengap
tolerances $10^{-10}$ and $10^{-6}$, and explicit deterministic $k$-NN tie
handling. The camera-ready code adopts the explicit theoretical lexicographic
tie rule. One $n=16$ $\varepsilon$-graph dataset failed the observed eigengap
check under the submitted policy and all alternatives; this leaves 399
evaluable datasets per policy and 2,394 evaluable comparisons overall.

\begin{table}[H]
\centering
\caption{Aggregate numerical-sensitivity audit. ``Set differences'' counts
comparisons with nonzero symmetric-difference probability mass between the two
computed selection sets. ``Discrete changes'' counts changes in evaluation
status, selected target, or eigengap validity.}
\label{tab:supp-numerical-audit}
\bigskip\small
\resizebox{\textwidth}{!}{\input{camera_numerical_audit_table.tex}}
\end{table}

Across the 2,394 evaluable comparisons, no alternative policy changed the
overall evaluation status, selected target, or eigengap validity. There were
484 comparisons with numerically nonzero symmetric-difference mass; every such
difference was confined to extremely narrow intervals adjacent to selection-set
boundaries. The maximum probability mass under the reference law was
$5.35\times10^{-16}$, and the maximum absolute $p$-value difference was
$1.10\times10^{-13}$. The only root-count change occurred for one balanced
separation-5 dataset under relaxed root merging (5,040 to 5,039 roots); it
merged a redundant root and changed neither the selection set's probability
mass nor the $p$-value.

We additionally selected 48 risk cases using small root spacing,
discriminants, graph margins, eigengaps, or truncation masses and recomputed
roots and compressed graph-state sequences at 100 decimal digits. The submitted
and long-double state sequences both matched the high-precision sequence in
48/48 cases. This high-precision check covers root and graph-state enumeration
only: spectral eigendecomposition and rounding remain float64 and are not
interval-enclosed. Consequently, the audit supplies empirical sensitivity
evidence rather than certified upper and lower truncation probabilities.

\section{Operational Usability and Power}
\label{supp:operational}

The submitted power files contain fixed attempts rather than only successful
replicates. Table~\ref{tab:supp-operational-power} therefore reports the
requested end-to-end quantities. Every cell had 500/500 usable attempts, so the
conditional rejection probability and the operational joint probability agree
in this grid. The exact finite-sample lower 95\% confidence bound on usability
in each cell is 0.993; this finite Monte Carlo result is not a universal
usability guarantee.

\begin{table}[H]
\centering
\caption{Operational usability and matched-rule power. Brackets are exact
binomial 95\% confidence intervals.}
\label{tab:supp-operational-power}
\bigskip\small
\resizebox{\textwidth}{!}{\input{camera_power_operational_table.tex}}
\end{table}

Table~\ref{tab:supp-operational-variants} gives the analogous attempt-based
quantities for the submitted graph/Laplacian robustness file. These runs used a
target-usable stopping rule, so the attempt totals are diagnostic denominators
rather than a prospectively fixed sample size. The only observed failures were
in the $n=16$ $\varepsilon$-graph cells: 99 eigengap failures before 2,000 usable
random-walk cases and 101 before 2,000 usable symmetric-normalized cases. All
other displayed cells were 2,000/2,000 usable.

\begin{table}[H]
\centering
\caption{Attempt-based operational quantities for graph and Laplacian variants.
Brackets are exact binomial 95\% confidence intervals.}
\label{tab:supp-operational-variants}
\bigskip\scriptsize
\resizebox{\textwidth}{!}{\input{camera_graph_operational_table.tex}}
\end{table}

\section{Feature-Wise Partial-Null Calibration with True Structure}
\label{supp:partial-null}

To complement the global-null experiments, we generated spherical Gaussian
data with $n=30,d=5$ and three balanced true groups. The first two coordinates
carried group separation $\delta$, while the fifth coordinate was pre-specified
for testing and had mean zero in every true group. We applied the feature-wise
matched-rule test to that null coordinate after spectral clustering. All 4,000
attempts were usable. Table~\ref{tab:supp-partial-null} shows calibration across
regimes ranging from restrictive truncation at $\delta=0$ to nearly vacuous
truncation at $\delta=5$.

\begin{table}[H]
\centering
\caption{Feature-wise partial-null calibration at nominal level .05. Brackets
are exact binomial 95\% confidence intervals.}
\label{tab:supp-partial-null}
\bigskip\small
\resizebox{\textwidth}{!}{\input{camera_partial_null_table.tex}}
\end{table}

Selective rejection rates were $0.050,0.048,0.043,$ and $0.054$ at
$\delta=0,2,4,5$, respectively. The corresponding naive rates were
$0.447,0.275,0.059,$ and $0.051$. A stable interval-mass helper and the submitted
direct formula differed by at most $3.70\times10^{-10}$ over the 4,000 attempts.
A pilot direct evaluation of the selection rule at 746 non-boundary points in 10
successful datasets found no mismatch with the enumerated intervals.

\section{Target-Rule Equivalence Sanity Check for the Reported \texorpdfstring{$K=3$}{K=3} Design}

Table~\ref{tab:supp-target-rule} reports a small null diagnostic for the main
$k$-NN, $L_{\mathrm{sym}}$, $K=3$ configuration. The reported target is the pair
of the two largest clusters. This table is not intended to show a power
difference among the first three conditioning events. In this $K=3$ setting,
once the two exact selected cluster index sets are fixed, the pair-appearance
and full-partition events coincide with the matched two-largest event: the
remaining cluster is the complement, and the selected pair retains the same
sizes. Thus the first three rows are expected to agree. The graph-conditioned
row is the stricter over-conditioned event used in the main paper's power
comparison. Larger-$K$ ablations can be generated with
\texttt{run\_target\_rule\_comparison.py}.

\begin{table}[H]
\centering
\caption{Target-rule equivalence sanity check under the global null
($n=32,d=5,K=3$, $k$-NN, $L_{\mathrm{sym}}$, $\sigma=1$).}
\label{tab:supp-target-rule}
\bigskip\small
\input{si_spectral_target_rule_table.tex}
\end{table}

\section{Robustness Matrix}

Table~\ref{tab:supp-variants} reports the graph-type, Laplacian, and
conditioning-level robustness matrix summarized in the main paper. The matched
rows condition on the two-largest-cluster target-selection rule; graph rows fix
the realized graph for the main $k$-NN, $L_{\mathrm{sym}}$ configuration.

\begin{table}[H]
\centering
\caption{Robustness matrix under the global null ($d=5$, $K=3$). Brackets give
exact binomial 95\% confidence intervals for the level $.05$ tail probability.}
\label{tab:supp-variants}
\bigskip\small
\resizebox{\textwidth}{!}{\input{si_spectral_robustness_table.tex}}
\end{table}

\section{Gaussian Data-Thinning Baseline}

Table~\ref{tab:supp-thinning} adds a simple independent-inference baseline for
the power experiment. For a thinning fraction $\rho=1/2$, draw
$E\sim N(0,\sigma^2 I)$ and set
\[
X^{\mathrm{sel}}=X+E,\qquad
X^{\mathrm{inf}}=X-E .
\]
Then $X^{\mathrm{sel}}$ and $X^{\mathrm{inf}}$ are independent, each with mean
$\mu$ and variance $2\sigma^2 I$. We run the same $k$-NN spectral clustering
pipeline on $X^{\mathrm{sel}}$, select the two largest clusters, and test their
mean difference on $X^{\mathrm{inf}}$ with the ordinary $\chi_d$ test using the
inflated scale $\sqrt{2}\sigma$. This baseline is valid because inference is
performed on data independent of the clustering step, but it answers a
different question: the selected clusters come from the randomized noisy copy,
not from the deterministic spectral clustering of the observed matrix $X$.
Power entries where thinning is larger should therefore be read as performance
for this randomized target, not as a direct power comparison for the same
reported clustering.

\begin{table}[H]
\centering
\caption{Rejection rates at level $.05$ for the matched selective test,
graph-conditioned selective test, and Gaussian data-thinning baseline. Each
entry uses $500$ usable replicates. The thinning baseline controls type-I error
but targets the randomized clustering of $X^{\mathrm{sel}}$.}
\label{tab:supp-thinning}
\bigskip\small
\input{si_spectral_thinning_baseline_table.tex}
\end{table}

\section{Scale-Sensitivity Check}

Table~\ref{tab:supp-scale} evaluates the same matched selection event under the
global null while changing only the scale used in the truncated-$\chi_d$ pivot.
The first row uses the true simulation scale and is the theorem-backed setting.
The plug-in and independent-pilot rows ignore scale-estimation uncertainty and
are included only as sensitivity checks; they should not be read as an
unknown-variance theorem. The last two rows show the expected directional effect
of a fixed 10\% scale misspecification.

\begin{table}[H]
\centering
\caption{Scale sensitivity under the global null
($n=32,d=5,K=3$, $k$-NN, $L_{\mathrm{sym}}$). Brackets give exact binomial
95\% confidence intervals for the level $.05$ tail probability.}
\label{tab:supp-scale}
\bigskip\small
\input{si_spectral_scale_sensitivity_table.tex}
\end{table}

\section{Distributional Stress Tests}

Table~\ref{tab:supp-stress} changes the null distribution while keeping the
same selective calculation and nominal $\sigma=1$ pivot. Only the spherical
Gaussian row is covered by the theorem. The standardized $t_5$ row is a
heavy-tailed non-Gaussian stress test; the correlated-Gaussian row keeps unit
marginal variances but violates the spherical feature covariance assumption.
These rows are included to show empirical sensitivity, not to claim validity
outside the model. The correlated-feature result is deliberately cautionary:
the $\chi_d$ pivot can be distorted when the feature covariance is not
proportional to the identity. This is the same caveat that applies to the PBMC
illustration: marginal standardization sets unit scale but does not make
correlated genes spherical.

\begin{table}[h]
\centering
\caption{Distributional stress tests under the global null
($n=32,d=5,K=3$, $k$-NN, $L_{\mathrm{sym}}$). Brackets give exact binomial
95\% confidence intervals for the level $.05$ tail probability.}
\label{tab:supp-stress}
\bigskip\small
\input{si_spectral_distribution_stress_table.tex}
\end{table}

\section{Fixed-Attempt Model-Scope Diagnostics}
\label{supp:misspecification}

Table~\ref{tab:supp-misspecification} extends the distributional stress analysis
with 500 fixed attempts per scenario. Only the first row is covered by the
spherical-Gaussian theorem. The other rows deliberately retain the same
selective calculation under row heteroskedasticity, row correlation, or feature
correlation and should be read only as scope diagnostics. All attempts were
usable. Row dependence was conservative in this particular design, whereas
feature correlation became increasingly anti-conservative; this does not
contradict the theorem because these rows violate its covariance assumptions.

\begin{table}[H]
\centering
\caption{Fixed-attempt model-scope diagnostics at nominal level .05. Brackets
are exact binomial 95\% confidence intervals. The KS column tests the full
$p$-value distribution against uniformity.}
\label{tab:supp-misspecification}
\bigskip\scriptsize
\resizebox{\textwidth}{!}{\input{camera_misspecification_table.tex}}
\end{table}

\section{Reproducibility Notes}

The code and computational artifacts described in this section are distributed with
the camera-ready PMLR supplementary package; the arXiv source package contains only
the paper and this textual supplement. The camera-ready Supplementary Material
contains the core implementation,
environment specification, fixed seeds, unit tests, quick and full reproduction
runners, scripts and raw and summary outputs for the originally submitted
experiments, and the manuscript sources. The result-to-script map in
\path{RESULTS_MAP.md} identifies the files used for each original table and
figure.

A dedicated rebuttal-results directory adds the compact raw and summary outputs
used for the camera-ready numerical audit, operational usability, partial-null,
misspecification, and component-runtime results. It also contains a
machine-readable reproducibility manifest recording the original execution
environment, seeds, commands, numerical policies, source hash, and output
checksums. The accompanying table-generation script validates the checked-in
summaries while regenerating the new LaTeX tables. The detailed interval
serialization file from the sensitivity audit is not required to rebuild the
paper tables and is omitted from the compact camera-ready package because of
its size; the included per-policy raw file, risk-case file, high-precision raw
file, summaries, and manifest contain the quantities reported here.

All theorem-level claims concern real arithmetic. The included outputs and
implementation use floating-point arithmetic; the numerical audit is empirical
sensitivity evidence and not interval certification. The 100-digit audit covers
root and graph-state sequences only, while eigendecomposition and rounding
remain float64. Likewise, the partial-null experiment is theorem-backed only for
its independent spherical-Gaussian design, and the non-spherical rows in
Table~\ref{tab:supp-misspecification} are scope diagnostics.

The principal original output map is:
\[
\begin{array}{lll}
\texttt{run\_primary\_calibration.py} &\to&
\texttt{si\_spectral\_primary\_calibration\_*},\\
\texttt{run\_robustness\_matrix.py} &\to&
\texttt{si\_spectral\_robustness\_*},\\
\texttt{run\_power\_curves.py} &\to&
\texttt{si\_spectral\_power\_curve\_*},\\
\texttt{run\_target\_rule\_comparison.py} &\to&
\texttt{si\_spectral\_target\_rule\_*},\\
\texttt{run\_thinning\_baseline.py} &\to&
\texttt{si\_spectral\_thinning\_baseline\_*},\\
\texttt{run\_scale\_sensitivity.py} &\to&
\texttt{si\_spectral\_scale\_sensitivity\_*},\\
\texttt{run\_distribution\_stress.py} &\to&
\texttt{si\_spectral\_distribution\_stress\_*},\\
\texttt{run\_scaling\_experiments.py} &\to&
\texttt{si\_spectral\_scaling\_*},\\
\texttt{run\_pbmc\_featurewise\_demo.py} &\to&
\texttt{pbmc\_featurewise\_*}.
\end{array}
\]

\section{Additional Scaling Diagnostics}

\begin{table}[H]
\centering
\caption{Scaling diagnostics for the finite-state homotopy computation under the
global null. Entries are averages over successful replicates; \texttt{Reps} is
the number of successful replicates used in the averages, and \texttt{Fail} is
the fraction of attempted replicates that failed the deterministic validity
checks. Peak memory is Python-side \texttt{tracemalloc} peak memory.}
\label{tab:supp-scaling}
\bigskip\small
\resizebox{\textwidth}{!}{\input{si_spectral_scaling_table.tex}}
\end{table}

\begin{table}[H]
\centering
\caption{Component timing for the current proof-of-concept implementation
(three fixed attempts per graph/$n$ cell). Percentages are fractions of total
wall-clock time in the recorded environment and are not hardware-independent
complexity constants.}
\label{tab:supp-runtime-components}
\bigskip\small
\resizebox{\textwidth}{!}{\input{camera_runtime_components_table.tex}}
\end{table}

\begin{figure}[H]
\centering
\includegraphics[width=\textwidth]{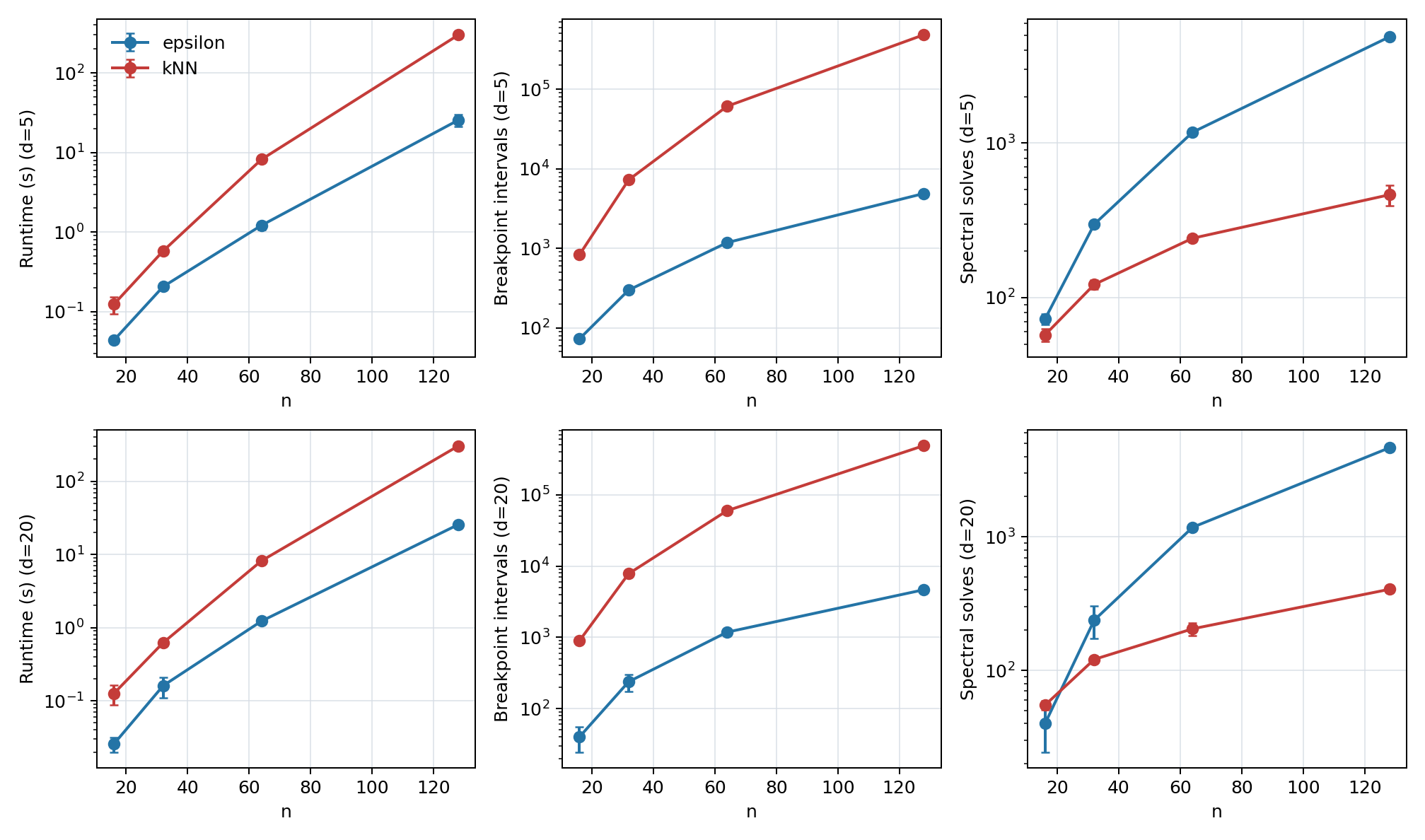}
\caption{Scaling of runtime, candidate breakpoint intervals, and spectral
solves for $d=5$ and $d=20$. The main paper reports the same diagnostics in
compressed form; Table~\ref{tab:supp-scaling} gives the numerical summaries.
This figure visualizes the separation between
candidate intervals and graph states requiring spectral recomputation.}
\label{fig:supp-scaling}
\end{figure}

\section{Working-Model PBMC Algorithmic Illustration}
\label{supp:pbmc}

The PBMC calculation is retained only to demonstrate the feature-wise homotopy
on a fixed preprocessed matrix. It is not a confirmatory biological analysis.
The protocol fixes a 16-gene panel, log-normalized library-size-adjusted counts,
marginal gene standardization, a fixed-seed subsample of 60 cells, $k=5$,
$K=3$, and the matched two-largest rule. It then sets a working scale
$\sigma=1$. This value is not a known sampling variance, and marginal
standardization does not remove gene--gene correlation. In light of
Table~\ref{tab:supp-misspecification}, the values below must not be interpreted
as scientifically valid PBMC $p$-values or marker discoveries.

\begin{table}[H]
\centering
\caption{Smallest working-model feature-wise values in the fixed 16-gene PBMC
panel. ``Adjusted'' means Bonferroni multiplication within this fixed panel.
The table is an algorithmic diagnostic, not confirmatory biological inference.}
\label{tab:supp-pbmc-working}
\bigskip\small
\input{pbmc_working_model_table.tex}
\end{table}

\begin{figure}[H]
\centering
\includegraphics[width=0.47\textwidth]{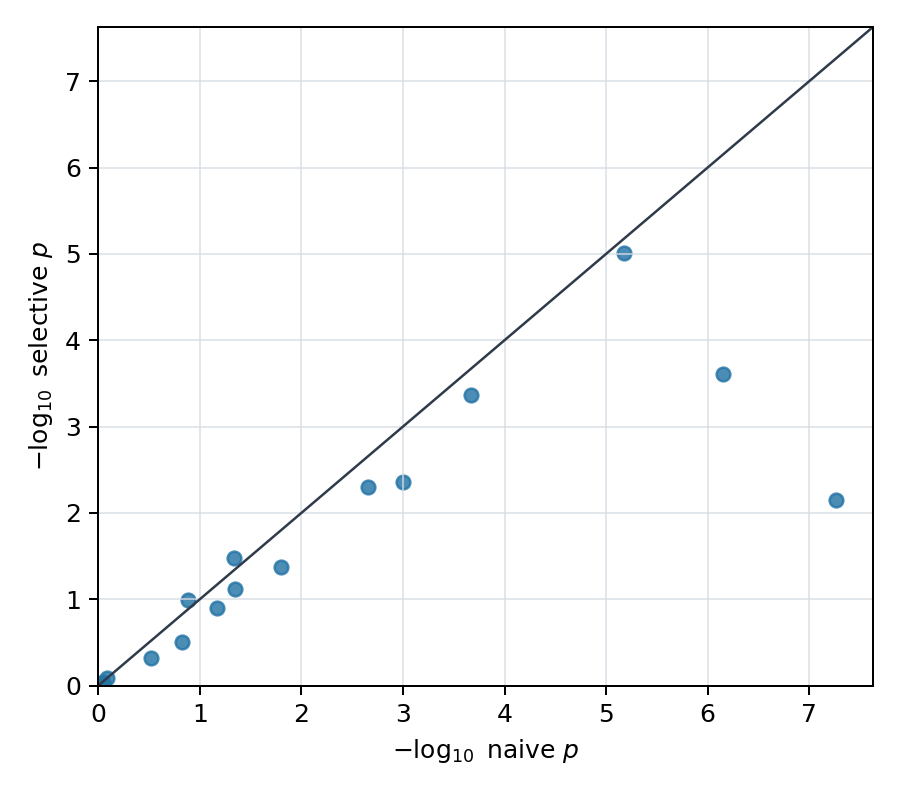}
\hfill
\includegraphics[width=0.47\textwidth]{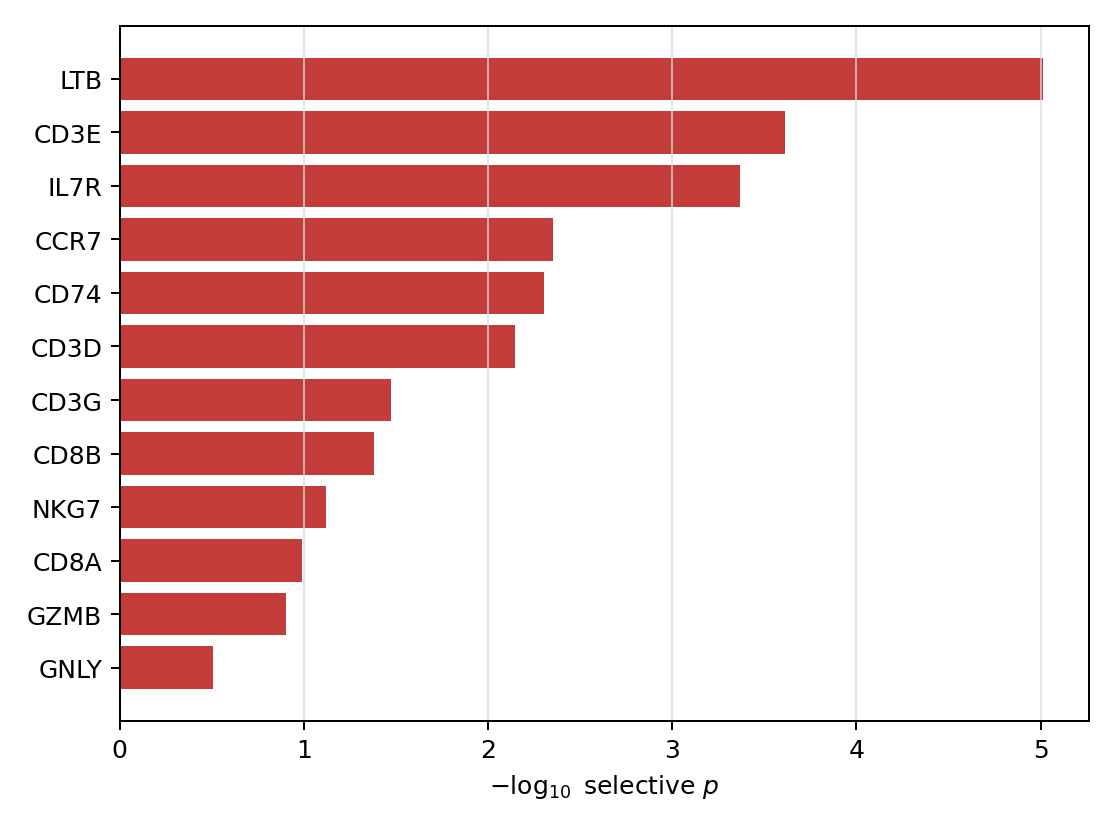}
\caption{Working-model PBMC diagnostics. Left: naive versus feature-wise
selective values. Right: genes ranked by the working-model selective value. The
plots illustrate numerical execution only; they do not establish marker
significance under a defensible PBMC sampling model.}
\label{fig:supp-pbmc}
\end{figure}

\end{document}

%% file: si_spectral_primary_calibration_table.tex
\begin{tabular}{lcccccc}
\toprule
Setting & $N$ & mean $p$ & $\Prob(p\le.01)$ & $\Prob(p\le.05)$ & $\Prob(p\le.10)$ & KS $p$\\
\midrule
current\_n18\_d5 & 5000 & 0.497 & 0.010 [0.008, 0.014] & 0.051 [0.045, 0.057] & 0.096 [0.088, 0.105] & 0.222\\
larger\_n32\_d5 & 5000 & 0.498 & 0.010 [0.008, 0.013] & 0.053 [0.047, 0.059] & 0.104 [0.096, 0.113] & 0.911\\
\bottomrule
\end{tabular}

%% file: camera_numerical_audit_table.tex
\begin{tabular}{lrrrrcc}
\toprule
Alternative policy & Attempted & Evaluable & Set differences & Discrete changes & Max. $|\Delta p|$ & Max. symmetric mass \\
\midrule
Stable quadratic formula & 400 & 399 & 121 & 0 & 1.10e-13 & 5.35e-16 \\
Long-double roots & 400 & 399 & 121 & 0 & 1.10e-13 & 5.35e-16 \\
Relaxed root merging & 400 & 399 & 121 & 0 & 1.10e-13 & 5.35e-16 \\
Eigengap tol. $10^{-10}$ & 400 & 399 & 0 & 0 & 0.00e+00 & 0.00e+00 \\
Eigengap tol. $10^{-6}$ & 400 & 399 & 0 & 0 & 0.00e+00 & 0.00e+00 \\
Explicit deterministic ties & 400 & 399 & 121 & 0 & 1.10e-13 & 5.35e-16 \\
\bottomrule
\end{tabular}

%% file: camera_power_operational_table.tex
\begin{tabular}{llcccc}
\toprule
Design & $\delta$ & Attempts & Usable & $\Pr(\mathrm{reject}\mid\mathrm{usable})$ & $\Pr(\mathrm{usable}\cap\mathrm{reject})$ \\
\midrule
Balanced & 0 & 500 & 500/500 & 0.046 [0.029, 0.068] & 0.046 [0.029, 0.068] \\
 & 1 & 500 & 500/500 & 0.052 [0.034, 0.075] & 0.052 [0.034, 0.075] \\
 & 2 & 500 & 500/500 & 0.158 [0.127, 0.193] & 0.158 [0.127, 0.193] \\
 & 3 & 500 & 500/500 & 0.448 [0.404, 0.493] & 0.448 [0.404, 0.493] \\
 & 4 & 500 & 500/500 & 0.832 [0.796, 0.864] & 0.832 [0.796, 0.864] \\
 & 5 & 500 & 500/500 & 0.952 [0.929, 0.969] & 0.952 [0.929, 0.969] \\
\addlinespace
Mildly unbalanced & 0 & 500 & 500/500 & 0.054 [0.036, 0.078] & 0.054 [0.036, 0.078] \\
 & 1 & 500 & 500/500 & 0.058 [0.039, 0.082] & 0.058 [0.039, 0.082] \\
 & 2 & 500 & 500/500 & 0.172 [0.140, 0.208] & 0.172 [0.140, 0.208] \\
 & 3 & 500 & 500/500 & 0.486 [0.441, 0.531] & 0.486 [0.441, 0.531] \\
 & 4 & 500 & 500/500 & 0.842 [0.807, 0.873] & 0.842 [0.807, 0.873] \\
 & 5 & 500 & 500/500 & 0.958 [0.937, 0.974] & 0.958 [0.937, 0.974] \\
\bottomrule
\end{tabular}

%% file: camera_graph_operational_table.tex
\begin{tabular}{ccclcccc}
\toprule
$n$ & Graph & Laplacian & Cond. & Attempts & $\Pr(\mathrm{usable})$ & $\Pr(\mathrm{reject}\mid\mathrm{usable})$ & $\Pr(\mathrm{usable}\cap\mathrm{reject})$ \\
\midrule
16 & $\varepsilon$ & $L_{\mathrm{rw}}$ & matched & 2099 & 0.953 [0.943, 0.962] & 0.050 [0.041, 0.060] & 0.048 [0.039, 0.058] \\
16 & $\varepsilon$ & $L_{\mathrm{sym}}$ & matched & 2101 & 0.952 [0.942, 0.961] & 0.056 [0.046, 0.066] & 0.053 [0.044, 0.063] \\
16 & $k$-NN & $L_{\mathrm{rw}}$ & matched & 2000 & 1.000 [0.998, 1.000] & 0.048 [0.039, 0.058] & 0.048 [0.039, 0.058] \\
16 & $k$-NN & $L_{\mathrm{sym}}$ & graph & 2000 & 1.000 [0.998, 1.000] & 0.044 [0.036, 0.054] & 0.044 [0.036, 0.054] \\
16 & $k$-NN & $L_{\mathrm{sym}}$ & matched & 2000 & 1.000 [0.998, 1.000] & 0.048 [0.039, 0.058] & 0.048 [0.039, 0.058] \\
32 & $\varepsilon$ & $L_{\mathrm{rw}}$ & matched & 2000 & 1.000 [0.998, 1.000] & 0.050 [0.041, 0.060] & 0.050 [0.041, 0.060] \\
32 & $\varepsilon$ & $L_{\mathrm{sym}}$ & matched & 2000 & 1.000 [0.998, 1.000] & 0.054 [0.045, 0.065] & 0.054 [0.045, 0.065] \\
32 & $k$-NN & $L_{\mathrm{rw}}$ & matched & 2000 & 1.000 [0.998, 1.000] & 0.053 [0.044, 0.064] & 0.053 [0.044, 0.064] \\
32 & $k$-NN & $L_{\mathrm{sym}}$ & graph & 2000 & 1.000 [0.998, 1.000] & 0.045 [0.037, 0.056] & 0.045 [0.037, 0.056] \\
32 & $k$-NN & $L_{\mathrm{sym}}$ & matched & 2000 & 1.000 [0.998, 1.000] & 0.043 [0.035, 0.053] & 0.043 [0.035, 0.053] \\
\bottomrule
\end{tabular}

%% file: camera_partial_null_table.tex
\begin{tabular}{cccccc}
\toprule
$\delta$ & Usable & Selective rejection & Operational joint & Naive rejection & Median truncation mass \\
\midrule
0 & 1000/1000 & 0.050 [0.037, 0.065] & 0.050 [0.037, 0.065] & 0.447 [0.416, 0.478] & 0.03007 \\
2 & 1000/1000 & 0.048 [0.036, 0.063] & 0.048 [0.036, 0.063] & 0.275 [0.248, 0.304] & 0.101 \\
4 & 1000/1000 & 0.043 [0.031, 0.057] & 0.043 [0.031, 0.057] & 0.059 [0.045, 0.075] & 0.9171 \\
5 & 1000/1000 & 0.054 [0.041, 0.070] & 0.054 [0.041, 0.070] & 0.051 [0.038, 0.067] & 1 \\
\bottomrule
\end{tabular}

%% file: si_spectral_target_rule_table.tex
\begin{tabular}{lccccc}
\toprule
Conditioning event & $N$ & mean $p$ & $\Prob(p\le.01)$ & $\Prob(p\le.05)$ & $\Prob(p\le.10)$\\
\midrule
Pair appears & 500 & 0.526 & 0.012 & 0.046 & 0.104\\
Two largest & 500 & 0.526 & 0.012 & 0.046 & 0.104\\
Full partition & 500 & 0.526 & 0.012 & 0.046 & 0.104\\
Realized graph & 500 & 0.518 & 0.008 & 0.042 & 0.096\\
\bottomrule
\end{tabular}

%% file: si_spectral_robustness_table.tex
\begin{tabular}{rlllcccc}
\toprule
$n$ & Graph & Laplacian & Cond. & $N$ & mean $p$ & $\Prob(p\le.05)$ & KS $p$\\
\midrule
16 & $k$-NN & $L_{\mathrm{sym}}$ & Matched & 2000 & 0.499 & 0.048 [0.039, 0.058] & 0.896\\
16 & $k$-NN & $L_{\mathrm{sym}}$ & Graph & 2000 & 0.499 & 0.044 [0.036, 0.054] & 0.802\\
16 & $k$-NN & $L_{\mathrm{rw}}$ & Matched & 2000 & 0.498 & 0.048 [0.039, 0.058] & 0.977\\
16 & $\varepsilon$ & $L_{\mathrm{sym}}$ & Matched & 2000 & 0.499 & 0.056 [0.046, 0.066] & 0.992\\
16 & $\varepsilon$ & $L_{\mathrm{rw}}$ & Matched & 2000 & 0.498 & 0.050 [0.041, 0.060] & 0.571\\
32 & $k$-NN & $L_{\mathrm{sym}}$ & Matched & 2000 & 0.502 & 0.043 [0.035, 0.053] & 0.817\\
32 & $k$-NN & $L_{\mathrm{sym}}$ & Graph & 2000 & 0.502 & 0.045 [0.037, 0.056] & 0.735\\
32 & $k$-NN & $L_{\mathrm{rw}}$ & Matched & 2000 & 0.499 & 0.053 [0.044, 0.064] & 0.776\\
32 & $\varepsilon$ & $L_{\mathrm{sym}}$ & Matched & 2000 & 0.504 & 0.054 [0.045, 0.065] & 0.724\\
32 & $\varepsilon$ & $L_{\mathrm{rw}}$ & Matched & 2000 & 0.491 & 0.050 [0.041, 0.060] & 0.080\\
\bottomrule
\end{tabular}

%% file: si_spectral_thinning_baseline_table.tex
\begin{tabular}{llrrrrrr}
\toprule
Design & Method & $\delta=0$ & $1$ & $2$ & $3$ & $4$ & $5$\\
\midrule
balanced & Matched selective & 0.046 & 0.052 & 0.158 & 0.448 & 0.832 & 0.952\\
balanced & Graph-conditioned & 0.050 & 0.042 & 0.068 & 0.134 & 0.418 & 0.660\\
balanced & Gaussian thinning & 0.046 & 0.086 & 0.234 & 0.728 & 0.964 & 0.998\\
\addlinespace
mildly unbalanced & Matched selective & 0.054 & 0.058 & 0.172 & 0.486 & 0.842 & 0.958\\
mildly unbalanced & Graph-conditioned & 0.060 & 0.038 & 0.076 & 0.216 & 0.512 & 0.770\\
mildly unbalanced & Gaussian thinning & 0.054 & 0.072 & 0.292 & 0.808 & 0.976 & 0.992\\
\bottomrule
\end{tabular}

%% file: si_spectral_scale_sensitivity_table.tex
\begin{tabular}{llcccc}
\toprule
Scale used in pivot & Theorem-backed? & $N$ & mean $p$ & $\Prob(p\le.05)$ & KS $p$\\
\midrule
True $\sigma$ & Yes & 500 & 0.497 & 0.056 [0.038, 0.080] & 0.765\\
Same-data plug-in $\hat\sigma$ & No & 500 & 0.496 & 0.052 [0.034, 0.075] & 0.701\\
Independent-pilot $\hat\sigma$ & No & 500 & 0.496 & 0.056 [0.038, 0.080] & 0.744\\
$0.9\sigma$ & No & 500 & 0.457 & 0.078 [0.056, 0.105] & 0.003\\
$1.1\sigma$ & No & 500 & 0.533 & 0.042 [0.026, 0.063] & 0.016\\
\bottomrule
\end{tabular}

%% file: si_spectral_distribution_stress_table.tex
\begin{tabular}{llcccc}
\toprule
Scenario & Theorem-backed? & $N$ & mean $p$ & $\Prob(p\le.05)$ & KS $p$\\
\midrule
Gaussian, spherical & Yes & 300 & 0.486 & 0.073 [0.047, 0.109] & 0.501\\
Standardized $t_5$ & No & 300 & 0.530 & 0.013 [0.004, 0.034] & 0.081\\
Gaussian, correlated features & No & 300 & 0.408 & 0.143 [0.106, 0.188] & 0.000\\
\bottomrule
\end{tabular}

%% file: camera_misspecification_table.tex
\begin{tabular}{lcccccc}
\toprule
Scenario & Theorem & Usable & Rejection & Operational joint & KS $p$ & Median truncation mass \\
\midrule
Spherical Gaussian & Yes & 500/500 & 0.052 [0.034, 0.075] & 0.052 & 0.809 & 6.57e-05 \\
Row heteroskedastic & No & 500/500 & 0.052 [0.034, 0.075] & 0.052 & 0.253 & 9.23e-05 \\
Row-correlated ($\rho=.2$) & No & 500/500 & 0.028 [0.015, 0.047] & 0.028 & 0.00112 & 0.000604 \\
Feature-correlated ($\rho=.1$) & No & 500/500 & 0.066 [0.046, 0.091] & 0.066 & 0.472 & 6.86e-05 \\
Feature-correlated ($\rho=.25$) & No & 500/500 & 0.068 [0.048, 0.094] & 0.068 & 0.0214 & 2.71e-05 \\
Feature-correlated ($\rho=.5$) & No & 500/500 & 0.144 [0.114, 0.178] & 0.144 & 4e-15 & 1.92e-05 \\
\bottomrule
\end{tabular}

%% file: si_spectral_scaling_table.tex
\begin{tabular}{lrrrrrrllll}
\toprule
Graph & $d$ & $n$ & Reps & Breakpoints & Intervals & Solves & Skipped & Time (s) & Peak MB & Fail \\
\midrule
eps & 5 & 16 & 3 & 72 & 73 & 73 & 0.0\% & 0.04 & 0.1 & 0.0\% \\
eps & 5 & 32 & 3 & 297 & 298 & 298 & 0.0\% & 0.21 & 0.2 & 0.0\% \\
eps & 5 & 64 & 3 & 1176 & 1177 & 1177 & 0.0\% & 1.22 & 0.7 & 0.0\% \\
eps & 5 & 128 & 3 & 4877 & 4878 & 4878 & 0.0\% & 25.53 & 2.5 & 0.0\% \\
eps & 20 & 16 & 2 & 39 & 40 & 40 & 0.0\% & 0.03 & 0.1 & 33.3\% \\
eps & 20 & 32 & 3 & 237 & 238 & 238 & 0.0\% & 0.16 & 0.3 & 0.0\% \\
eps & 20 & 64 & 3 & 1179 & 1180 & 1180 & 0.0\% & 1.23 & 1.2 & 0.0\% \\
eps & 20 & 128 & 3 & 4646 & 4647 & 4647 & 0.0\% & 25.56 & 4.5 & 0.0\% \\
knn & 5 & 16 & 3 & 829 & 830 & 57 & 93.1\% & 0.12 & 0.1 & 0.0\% \\
knn & 5 & 32 & 3 & 7242 & 7243 & 122 & 98.3\% & 0.58 & 0.7 & 0.0\% \\
knn & 5 & 64 & 3 & 60892 & 60893 & 242 & 99.6\% & 8.19 & 4.6 & 0.0\% \\
knn & 5 & 128 & 3 & 482753 & 482754 & 463 & 99.9\% & 299.73 & 24.9 & 0.0\% \\
knn & 20 & 16 & 3 & 894 & 895 & 55 & 93.8\% & 0.13 & 0.1 & 0.0\% \\
knn & 20 & 32 & 3 & 7809 & 7810 & 121 & 98.4\% & 0.62 & 0.6 & 0.0\% \\
knn & 20 & 64 & 3 & 59744 & 59745 & 205 & 99.7\% & 8.18 & 4.2 & 0.0\% \\
knn & 20 & 128 & 3 & 488983 & 488984 & 405 & 99.9\% & 302.92 & 26.5 & 0.0\% \\
\bottomrule
\end{tabular}

%% file: camera_runtime_components_table.tex
\begin{tabular}{ccccccc}
\toprule
Graph & $n$ & Candidate intervals & Spectral solves & Total sec. & Graph walk & Eigen/rounding \\
\midrule
$\varepsilon$ & 32 & 273 & 273.3 & 0.06 & 9.4\% & 87.8\% \\
$\varepsilon$ & 64 & 1,186 & 1,186.0 & 0.56 & 6.8\% & 92.8\% \\
$\varepsilon$ & 128 & 4,989 & 4,989.0 & 6.54 & 5.5\% & 94.5\% \\
$k$-NN & 32 & 6,890 & 107.3 & 0.18 & 83.9\% & 13.9\% \\
$k$-NN & 64 & 59,083 & 240.3 & 5.89 & 97.3\% & 2.4\% \\
$k$-NN & 128 & 497,839 & 482.0 & 244.59 & 99.6\% & 0.3\% \\
\bottomrule
\end{tabular}

%% file: pbmc_working_model_table.tex
\begin{tabular}{lrrrr}
\toprule
Gene & Mean diff. & Naive value & Selective value & Adjusted selective value \\
\midrule
LTB & 1.27 & 6.61e-06 & 9.72e-06 & 1.56e-04 \\
CD3E & 1.40 & 6.98e-07 & 2.44e-04 & 3.90e-03 \\
IL7R & 1.05 & 2.15e-04 & 4.29e-04 & 6.87e-03 \\
CCR7 & 0.93 & 9.99e-04 & 4.44e-03 & 7.10e-02 \\
CD74 & -0.87 & 2.20e-03 & 4.95e-03 & 7.93e-02 \\
CD3D & 1.54 & 5.41e-08 & 7.14e-03 & 1.14e-01 \\
CD3G & 0.56 & 4.57e-02 & 3.36e-02 & 5.38e-01 \\
CD8B & 0.68 & 1.59e-02 & 4.18e-02 & 6.69e-01 \\
NKG7 & -0.57 & 4.56e-02 & 7.57e-02 & 1.00e+00 \\
CD8A & 0.43 & 1.31e-01 & 1.03e-01 & 1.00e+00 \\
GZMB & -0.52 & 6.76e-02 & 1.25e-01 & 1.00e+00 \\
GNLY & -0.41 & 1.50e-01 & 3.11e-01 & 1.00e+00 \\
\bottomrule
\end{tabular}